\documentclass[preprint,12pt]{elsarticle}

\usepackage{amssymb}
\usepackage{amsmath}

\usepackage{algorithm}
\usepackage{algorithmic}
\usepackage{makecell}
\usepackage{multirow}
\usepackage{multicol}
\usepackage{wrapfig}
\usepackage{flushend}
\usepackage{graphicx}
\usepackage{tabularx}
\usepackage[a4paper,margin=1in]{geometry}
\usepackage{array}    
\usepackage{fontawesome} 
\usepackage{tabularray} 
\usepackage{xcolor}
\usepackage{booktabs}
\usepackage{caption}
\usepackage{pifont}
\newcommand{\cmark}{\ding{51}} 
\newcommand{\xmark}{\ding{55}} 
\usepackage{amsthm}
\newtheorem{remark}{Remark}
\usepackage{bm}

\journal{Information Sciences}

\begin{document}

\begin{frontmatter}



\title{{Federated Hierarchical Clustering with Automatic Selection of Optimal Cluster Numbers}}


\author[1]{Yue Zhang}
\ead{zhangyue@gpnu.edu.cn}

\author[1]{Chuanlong Qiu}
\ead{q1870088463@gmail.com}

\author[1]{Xinfa Liao}
\ead{L593486501@gmail.com}

\author[2,3]{Yiqun Zhang\corref{cor1}}
\ead{yqzhang@comp.hkbu.edu.hk}

\cortext[cor1]{Corresponding author.}

\affiliation[1]{organization={School of Computer Science, Guangdong Polytechnic Normal University},
            city={Guangzhou},
            country={China}}

\affiliation[2]{organization={School of Computer Science and Technology, Guangdong University of Technology},
            city={Guangzhou},
            country={China}}

\affiliation[3]{organization={Department of Computer Science, Hong Kong Baptist University},
            city={Hong Kong SAR},
            country={China}}

\begin{abstract}
{Federated Clustering (FC) is an emerging and promising solution in exploring data distribution patterns from distributed and privacy-protected data in an unsupervised manner. Existing FC methods implicitly rely on the assumption that clients are with a known number of uniformly sized clusters. However, the true number of clusters is typically unknown, and cluster sizes are naturally imbalanced in real scenarios. Furthermore, the privacy-preserving transmission constraints in federated learning inevitably reduce usable information, making the development of robust and accurate FC extremely challenging. Accordingly, we propose a novel FC framework named Fed-$k^*$-HC, which can automatically determine an optimal number of clusters $k^*$ based on the data distribution explored through hierarchical clustering. To obtain the global data distribution for $k^*$ determination, we let each client generate micro-subclusters. Their prototypes are then uploaded to the server for hierarchical merging. The density-based merging design allows exploring clusters of varying sizes and shapes, and the progressive merging process can self-terminate according to the neighboring relationships among the prototypes to determine $k^*$. Extensive experiments on diverse datasets demonstrate the FC capability of the proposed Fed-$k^*$-HC in accurately exploring a proper number of clusters.}

\end{abstract}

\begin{highlights}
\item Tackles imbalanced cluster distribution in one-shot federated clustering. 
\item Fine partitioning and hierarchical merging mitigate the “uniform effect”.
\item Automatically identifies the optimal number of clusters in federated learning.
\end{highlights}


\end{frontmatter}



\section{Introduction}\label{sec1}
Federated learning \cite{konecny2016federated2} is a distributed learning method aimed at protecting data privacy during collaborative learning between devices \cite{zhang2021survey}. It has been extensively applied in a variety of practical scenarios, including medical diagnosis \cite{nazir2023federated}, financial fraud detection \cite{mustafa2024federated}, industrial manufacturing \cite{koley2024critically}, etc., allowing devices to train models locally and only share model parameters to build a global model \cite{yang2019federated}, thereby protecting data privacy \cite{ALHUTHAIFI2023833, ZeinabPPFL2025}. Since most real-world datasets are unlabeled, Federated Clustering (FC) has emerged as a critical tool for data analysis in federated environments \cite{issuesINfl_Bhanbhro2024IssuesIF}. However, the original unsupervised challenges are further amplified by privacy-preserving constraints in the federated learning scenarios. That is, the inherent need for detailed sample information in unsupervised learning fundamentally conflicts with the privacy-preserving restrictions on data transmission of federated learning, which mutually bring difficulties to FC \cite{Lubana2022OrchestraUF}.

To address the insufficient clustering information caused by the federated transmission constraints, existing FC approaches can be categorized into two main types: 1) raw data encryption methods and 2) data distribution sketch methods. The former type involves encrypting raw data on clients to enable secure uploading for informative server-end distribution analysis \cite{Li2023OnTP}. However, these complex encryption algorithms may substantially increase computational overhead and communication costs \cite{Hegde2021SoKEP, abood2018survey}. In edge-dominant federated learning scenarios, the applicability of such methods can be severely constrained. Moreover, it is often difficult to perform effective statistical analysis or modeling directly on encrypted data, which prevents the server from effectively learning comprehensive cluster distributions. In contrast, the latter type of approach performs local data analysis and representation encoding at the client side before transmitting only the extracted information to the server, thereby effectively safeguarding raw data from exposure. A representative FC method in this stream \cite{Zhu2023F3KMFF} only transmits descriptive parameters of the learned clusters, e.g., cluster centroids, or alternatively, partially synthetic surrogate data to be uploaded to the server to represent the raw data \cite{pan2023machine}. Accordingly, they can significantly mitigate privacy risks while preserving clustering performance across distributed heterogeneous data sources. Nevertheless, almost all these methods rely on strong assumptions about cluster distributions: 1) the clusters are of uniform sizes, and 2) a predetermined proper number of clusters $k$ is given, which rarely holds in real scenarios. Consequently, these methods either fail to capture naturally imbalanced clusters by forcibly assigning samples to equally sized groups, or impose uniform cluster counts across all clients despite their fundamentally heterogeneous data distributions.

\begin{table}[!t]
\caption{Comparison of key characteristics of existing FC methods.}
\centering
\begin{tabular}{l|cccc}
\toprule
 Method &  Non-IID & Imbalance & Auto-$k^*$ & One-shot         \\
\midrule
 KFed \cite{dennis2021heterogeneity} & \cmark       & \xmark       & \xmark  & \cmark  \\
 MUFC \cite{pan2023machine} & \cmark       & \xmark  & \xmark & \cmark  \\
 F3KM \cite{ Zhu2023F3KMFF} & \cmark       & \cmark       & \xmark & \xmark   \\
 Orchestra \cite{ Lubana2022OrchestraUF} & \cmark       & \xmark       & \xmark & \xmark  \\
 FedET \cite{liu2023fedet} & \cmark       & \cmark       & \xmark & \xmark   \\
 FedGT \cite{zhang2024fedgt} & \cmark       & \cmark       & \xmark & \xmark   \\
\midrule
 \textbf{Fed-$k^*$-HC(ours)} & \cmark       & \cmark       & \cmark & \cmark  \\
\bottomrule
\end{tabular}
\label{methodInfo}
\end{table}

To achieve more robust representation learning for complex distributions under the federated learning scenario, deep federated models incorporating Transformer architectures \cite{liu2023fedet} and graph representation learning \cite{zhang2024fedgt} have been progressively developed. Although they demonstrate attractive potential in exploring imbalanced distributions, their supervised design prevents them from being applied to fully unsupervised FC tasks. A summary of representative FC approaches is organized in Table \ref{methodInfo} from several key aspects, including the capabilities to handle datasets with complex distributions, i.e., non-IID and imbalanced data with a unknown number of optimal clusters $k^*$ to be automatically determined (Auto-$k^*$), and whether the learning is performed in an advanced one-shot manner to mitigate privacy leakage risks. As our previous analysis demonstrates, achieving a balance between privacy preservation and unsupervised cluster distribution learning is inherently challenging. Consequently, as illustrated in Table \ref{methodInfo}, existing methods remain fundamentally limited in simultaneously exploring complex cluster distributions while providing the advanced one-shot privacy protection with fewer communication rounds.

This paper, therefore, proposes a novel FC method under the one-shot \underline{\textbf{Fed}}erated learning framework to automatically determine \underline{\textit{$k^*$}} based on informative data distribution explored through \underline{\textbf{H}}ierarchical \underline{\textbf{C}}lustering, named Fed-$k^*$-HC. Unlike existing FC methods with strong and ideal cluster distribution assumptions, our approach adopts a novel paradigm to obtain micro clusters on the clients to finely describe the divergent distributions of clients, and then merge them represented by their uploaded prototypes on the server to form a comprehensive global distribution of the distributed data. First, data is clustered into excessive micro-subclusters on the clients, approximating the cluster distributions with many fine-grained and compact subclusters without biasing certain cluster shapes and distribution types. Then, a corresponding number of synthetic data points, generated based on the features of the subclusters, are uploaded to the server to represent the data from clients. The server processes them by sequentially merging the current most similar subclusters across different clients, and the appropriate number of clusters $k^*$ is automatically determined when the closely connected neighbors are all merged. Such a gradual merging process helps to prevent the premature merging or neglecting of smaller clusters. Furthermore, the hierarchical clustering paradigm mitigates the common ``uniform effect'' seen in partition-based clustering algorithms, thus allowing the precise detection of imbalanced clusters described by the Non-IID data from the clients. It turns out that the more secure one-shot FC and the more informative hierarchical fusion facilitate an effective complementary, elegantly circumventing the fundamental security-information trade-off in FC. Compared to existing FC methods, the proposed Fed-$k^*$-HC demonstrates superior performance in accurately explore a proper number of imbalanced clusters, which is practical in the analysis of complex real datasets. The main contributions are summarized as follows:
\begin{itemize}
\item \textbf{A new federated clustering paradigm: }This paper addresses the widely prevalent yet poorly solved issue of federated clustering with imbalanced data, providing an effective federated clustering paradigm that lays the foundation for future research. 
\item \textbf{Fine-partition and hierarchical merging mechanism: }We propose partitioning the local data of clients into tiny subclusters and performing hierarchical merging based on the regenerated data information uploaded to the server. As a result, our algorithm achieves better performance in clustering imbalanced data.
\item \textbf{Automatic $k^*$ determination under FL setting: }Unlike existing federated clustering algorithms, our proposed method can automatically determine the appropriate $k^*$, significantly eliminating the assumptions regarding the clustering distribution on the clients and granting them greater flexibility to explore clustering distributions that are closer to the actual data.
\end{itemize}

\section{Related Works}\label{sec2}

\subsection{Federated Clustering}
Current research on clustering tasks in federated learning primarily includes the following aspects. 

\textbf{Clustered Federated Learning: }Considering potential confusion regarding terminology, this category of research is first introduced, and the differences between them and the present study are highlighted. The field of Clustered Federated Learning explores how clustering can assist in supervised federated learning to handle non-IID data more effectively \cite{briggs2020federated}. These studies achieve this by clustering clients, focusing on completing subsequent supervised learning tasks, with most employing iterative or centralized clustering schemes \cite{sattler2020clustered}.

\textbf{Federated clustering:} Our research focuses on this category of methods, emphasizing the identification of the distribution of global clusters in distributed data without sharing raw data. Currently, researchers have attempted to extend the $k$-means algorithm to federated environments, proposing to update global centroids using a global averaging function. However, in scenarios with highly skewed data distributions or high-dimensional datasets (e.g., FEMNIST), the clustering performance tends to fluctuate significantly \cite{kumar2020federated}. Additionally, a fuzzy version of this approach uses fuzzy assignments as weights. Nevertheless, sharing fuzzy membership information may pose risks of privacy leakage \cite{pedrycz2021federated}. Although this method has not yet been fully explored, it has shown potential in some experiments for identifying reasonable clusters. Li et al. \cite{li2022secure} proposed a federated clustering approach from another perspective, where local data is encrypted on the client side before being sent to the server for clustering. However, the method incurs high computational and communication overhead in large-scale federated settings, limiting its scalability and efficiency.

Further research has highlighted the potential of one-shot clustering as a simple preprocessing step to enable personalized federated learning \cite{dennis2021heterogeneity}. Although federated learning approaches may increase communication overhead to some extent \cite{li2020federated}, this method \cite{dennis2021heterogeneity} achieves performance comparable to or better than new iterative methods without requiring multiple rounds of communication \cite{ghosh2020efficient}. Building on $k$-means, Pan et al. \cite{pan2023machine} were the first to introduce the problem of machine unlearning in federated clustering, designing an efficient federated clustering framework that integrates secure compressed multiset aggregation (SCMA) technology. Zhang et al. \cite{zhang2025asynchronous} proposed an innovative asynchronous federated clustering learning (AFCL) method that addresses the challenges of asynchronous client communication and unknown cluster numbers, providing a novel solution for privacy-preserving distributed clustering. This framework enables efficient data aggregation and clustering while preserving user privacy and supports fast deletion requests of user data. The above $k$-means-based federated clustering methods are primarily focused on theoretical research. On the practical side, Lu et al. \cite{LU2023105714} developed a federated clustering method based on the global weighted $k$-means algorithm for analyzing driving behavior data while preserving user privacy, aiming to identify different driving styles (e.g., aggressive, steady, or calm). Wang et al. \cite{Wang2022FederatedCF} applied the federated $k$-means method to smart meter data analysis, extracting typical power consumption patterns from the data to help retailers better understand consumer behavior. This, in turn, enables personalized pricing designs, demand response management, and other diversified services. In addition to the $k$-means-based methods, Xie et al. \cite{xie2023fed} conducted research on federated clustering for high-dimensional data and proposed a federated subspace clustering method to address the limitations of 
$k$-means clustering in high-dimensional spaces.

Although many federated clustering methods remain based on $k$-means and its variants, recent research has begun to introduce Transformer-based architectures to address more complex and non-IID federated data. For instance, FedET \cite{liu2023fedet} employs an enhanced Transformer architecture and integrates incremental class distillation to mitigate issues related to class imbalance and continuous learning; FedGT \cite{zhang2024fedgt} extends Transformer models to graph-structured data, enabling subgraph-level clustering and federated aggregation. However, these methods have several limitations. First, the Transformer model itself has a large number of parameters and high computational overhead, making it challenging to deploy on resource-constrained clients. Second, the training process is sensitive to data quality and label distribution, which can lead to overfitting or inadequate generalization. Furthermore, Transformers exhibit poor stability in small-sample and high-noise environments, making training convergence difficult. Lastly, in a federated environment, model synchronization and alignment become more challenging due to the data heterogeneity across clients. While graph-based methods like FedGT have enhanced structural modeling capabilities, their aggregation efficiency and clustering accuracy in large-scale, sparse, or dynamic graph data remain suboptimal, limiting their applicability. 

These studies explore clustering in federated learning from different angles, providing various methods and insights to address clustering challenges in distributed computing environments.

\subsection{Imbalanced Clustering}
Regarding the issue of imbalanced clustering, researchers have primarily focused on the imbalance of data classes, as demonstrated in Lin et al.'s study \cite{lin2017clustering}. However, the scenario of imbalanced clustering with unequal sample sizes within clusters has not received sufficient attention. Nevertheless, some studies have explored clustering on imbalanced data. For instance, Xiong et al. \cite{xiong2009k} examined the performance of $k$-means on imbalanced data, introducing the concept of the ``uniform effect'' and proposing the coefficient of variation as an evaluation metric. Liang et al. \cite{liang2012k} developed an algorithm based on fuzzy $k$-means clustering, which employs a three-stage process for clustering imbalanced data. However, this method faces challenges such as predefined prototype numbers and manual selection of the number of clusters. Additionally, some studies have used ensemble clustering or spectral clustering methods to address clustering issues in imbalanced data, but these approaches still fail to handle imbalanced clusters effectively.

In this context, Zhao et al. \cite{10.1007/978-981-99-8435-0_3} developed an unsupervised concept drift detection approach that can effectively explore imbalanced clusters and monitor their drifts. Most recently, Zhang et al. \cite{8423698} further proposed an imbalanced clustering algorithm for efficient and automatic adaptation of streaming data cluster distribution. Although recent advanced imbalanced clustering methods have been developed, the imbalanced distribution has not been explicitly considered in the field of FC.

\subsection{Selecting the Number of Clusters in Clustering}
In the existing research on federated clustering, the selection of the appropriate number of clusters has not been explored in much detail; most approaches treat the number of clusters as an input to the algorithm, implying that it is considered a known condition. However, in practical clustering tasks, the number of clusters is not always known. Therefore, based on some existing studies, the automatic determination of an appropriate number of clusters is explored. Some relevant works that will be discussed are introduced here. Fritzke \cite{Fritzke1994GCS} proposed an unsupervised neural network model called ``Growing Cell Structures'' (GCS), which can automatically adjust the size and structure of the network to adapt to data distribution. GCS enhances clustering performance by dynamically adding or removing nodes, enabling more flexible self-organizing learning. Zhu et al. \cite{Zhu2016NN} introduced a Natural Neighbor algorithm (NaN) for data classification and anomaly detection, which adaptively constructs a neighborhood graph to replace the fixed parameter $k$ in traditional 
``$k$-Nearest Neighbor \cite{knn}'' algorithms. This method better reflects the local characteristics of the data, thereby improving classification accuracy and the detection of outliers.

\section{Method}\label{sec3}
\begin{figure}[t]
\centerline{\includegraphics[width=1\textwidth]{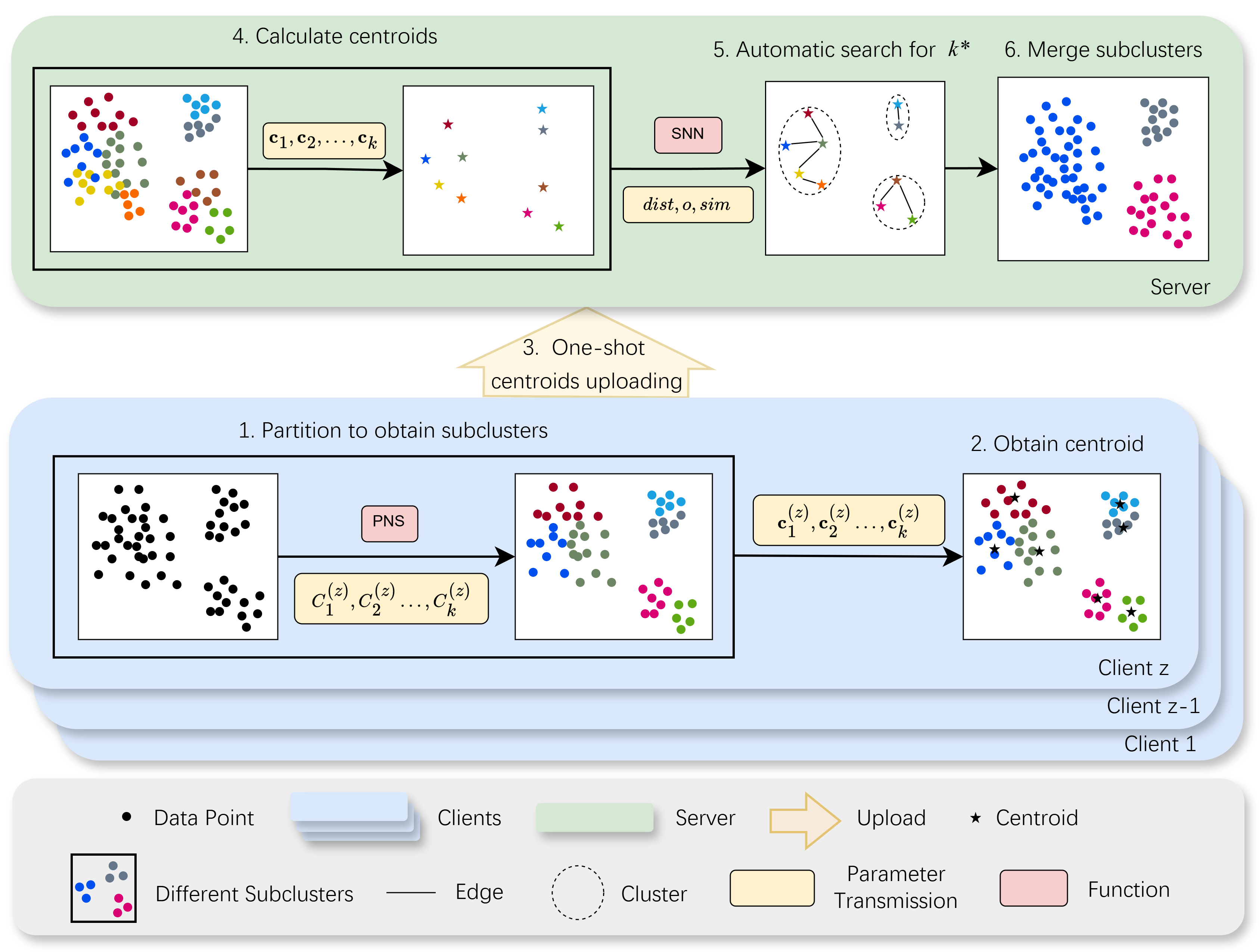}}
\caption{Overview of the proposed Fed-$k^*$-HC framework. Each client first applies the SNP algorithm to partition its local data (black dots) into multiple micro-subclusters (colored dots) and computes the centroid of each subcluster (black stars). These centroids are then uploaded to the server. The server aggregates all client subclusters (colored dots), reinitializes global centroids (colored stars), and performs clustering using corresponding neighborhood-based algorithms. The number of clusters $k^*$ is automatically estimated via a neighborhood-based method, without requiring manual specification.}\label{overview}
\end{figure}

In this section, we first define the FC problem, and then present the proposed Fed-$k^*$-HC algorithm composed of two key technical components: 1) Client-Side Automated Micro-Partitioning, and 2) Server-Side Hierarchical Merging. Frequently used notations are summarized in Table \ref{notations}, and the overview of Fed-$k^*$-HC is shown in Figure \ref{overview}. The FC task is to cluster imbalanced data distributed across multiple devices under the federated learning configuration. The characteristics of federated learning prohibit centralized access to raw data, leading to significant differences in data distribution and sample sizes among clients. This data heterogeneity and imbalance can severely affect the accuracy and reliability of clustering results. For example, in a federated mobile application, some users generate frequent data (e.g., daily fitness tracking), while others contribute only occasionally. Behavioral patterns also vary across regions. Ignoring such an imbalance may cause clusters to reflect only dominant user groups, overlooking minority patterns. Therefore, our primary goal is to ensure the effective protection of user privacy without transmitting raw sample data while obtaining a proper number of potentially imbalanced global clusters on the server.

\begin{table}[!t]
    \centering
    \caption{Main notations and abbreviations}.
    \begin{tabular}{l|l}
        \toprule
        Notation \&  \\ Abbreviation & \multirow{2}{*}[2.5ex]{Description}  \\ \midrule
        $C_g^{(z)}$&The set of data points in the $g$-th subcluster on client $z$.\\ 
         $d_{C_i, C_j}$& The special distance between two subclusters $C_i, C_j$.
\\
        $o_{C_i, C_j}$& The overlap degree between two subclusters $C_i, C_j$.\\
 $K$& The ground-truth number of clusters in dataset \textbf{X}.  \\
 $k^*$& The number of automatically determined clusters.\\
 $b$&The $b$-nearest neighbor parameter.  \\
 $\mathbf{m}_{c}$&The winner seed point.\\
 $\mathbf{m}_{j}$&The opponent seed point.\\
 $\alpha_{c}$&The learning rate.\\
 $\beta_{j}$&The opponent penalty coefficient.\\
 ${\delta_j}$&The density gap of each subcluster.\\
 ${\bar{d}_j^{(z)}}$&The averange distance between all point pairs in ${C_g^{(z)}}$.\\

 $\text{NN}_b(\mathbf{x}_i)$& The set of $b$-nearest neighbors of point $\mathbf{x}_i$.  
\\
 
$\text{NN}_b^{(m)}(\mathbf{x}_i)$& The $m$-th nearest neighbor of $\mathbf{x}_i$ within $\text{NN}_b(\mathbf{x}_i)$.  
\\
 SNP& Selection of Number of Prototypes  \\
 SNS& Selection of Number of Clusters  \\
 LNN& Loose Natural Neighbors  \\
 SNN& Strict Natural Neighbors  \\
  \bottomrule
    \end{tabular}
    
    \label{notations}
\end{table}

To achieve this goal, a hierarchical clustering paradigm is adapted to the federated learning settings. It employs a bottom-up approach for subcluster merging, which not only ensures that small clusters are effectively identified and processed but also significantly alleviates the issues caused by centroid shift in existing partition-based clustering methods. Through this approach, the true structure of the data is more accurately reflected, thereby improving the overall performance and adaptability of clustering. Our method consists of local clustering on the client, determination of the number of clusters, and global clustering on the server. On each client, the data is partitioned into several subclusters, and then substitute data, generated using the multivariate normal distribution, is created in the same quantity as the original data and uploaded to the server. This approach effectively approximates the original data distribution while preserving privacy, as each micro subcluster can be viewed as a non-partitional minimal distribution unit. Such micro-units can together approximate arbitrary complex cluster distributions without biasing certain cluster shapes and distribution types, and can capture key statistical properties (i.e., their mean and covariance) to be uploaded to the server without leaking raw data samples. On the server, based on the received data, an adaptive merging strategy is used to determine the appropriate number of clusters, ultimately producing the global clustering result. 

\subsection{Client-Side Automated Micro-Partitioning}

Since the raw data from the clients cannot be directly uploaded to the server in federated learning, the data is first partitioned into several micro subclusters on each client. These subclusters are then processed and uploaded to the server for further handling. Assuming a federated setting with $Z$ clients, indexed by $z \in \{1,2, \ldots, Z\}$. Each client $z$ holds a local dataset denoted by $\textbf{X}^{(z)}=\{{\mathbf{x}_i^{(z)}} \in \mathbb{R}^d \mid i=1,2, \ldots, n^{(z)}\}$, where $\textbf{x}_i^{(z)}$ is a $d$-dimensional feature vector, and $n^{(z)}$ represents the number of data samples on client $z$. The entire dataset is the union of all local datasets, and the total number of samples is given by $\sum_{z=1}^Z n^{(z)}=n$. The local dataset $\textbf{X}^{(z)}$ can be further partitioned into $k$ subclusters $\{{C_1^{(z)}, C_2^{(z)}, \ldots, C_k^{(z)}}\}$, where $C_g^{(z)} \subseteq \textbf{X}^{(z)}$ denotes the $g$-th subcluster on client $z$, for $g \in \{1,2, \ldots, k\}$. Various existing methods can be employed to partition $\textbf{X}^{(z)}$, including but not limited to DBSCAN \cite{schubert2017dbscan}, Affinity Propagation \cite{frey2007clustering}, and Mean Shift \cite{carreira2015review}. Through these methods, a series of subclusters $\{{C_1^{(z)}, C_2^{(z)}, \ldots, C_{k}^{(z)}}\}$ are obtained, along with their corresponding centroids $\{{\mathbf{c}_1^{(z)}, \mathbf{c}_2^{(z)}, \ldots, \mathbf{c}_k^{(z)}}\}$, where $\mathbf{c}_g^{(z)}$ denotes the centroid of subcluster $C_g^{(z)}$, and $g \in \{1,2, \ldots, k\}$.

In this process, if the sizes of the obtained subclusters are similar, the similarity between subclusters from different clusters will be relatively small, which can improve the merging performance of subclusters on the server. Therefore, we adopt a competitive learning method called Selection of Number of Prototypes (SNP) \cite{lu2019self}, which can automatically determine an appropriate number of prototypes to represent the data. It uses a multi-prototype cluster mechanism, where each cluster can be represented by one or more subclusters. Specifically, for each data point {$\mathbf{x}_i^{(z)}$}, the nearest seed point is selected as the ``winner'' and its position is updated according to: 
\begin{equation}
\mathbf{m}_c = \mathbf{m}_c + \alpha_c (\mathbf{x}_i^{(z)} - \mathbf{m}_c),
\end{equation}
where {$\mathbf{m}_c$} denotes the winner seed point and {$ \alpha_c$} is the learning rate associated with the winner. For each opponent seed point {$\mathbf{m}_j$}, its position is updated based on the distance to the winner and a penalty coefficient {$\beta_j$}. The update rule is given by:
\begin{equation}
\mathbf{m}_j = \mathbf{m}_j - \eta \beta_j \alpha_c (\mathbf{x}_i^{(z)} - \mathbf{m}_c),
\end{equation}
where {$\eta$} is a parameter controlling the strength of the penalty, and {$\beta_j$} is the opponent penalty coefficient. To determine whether new seed points should be introduced, the density gap {$\delta_j$} of each sub-cluster is computed:
\begin{equation}
\delta_j = \max_{\mathbf{x}_i^{(z)} \in C_j^{(z)}} \min_{\substack{\mathbf{x}_j^{(z)} \in C_j^{(z)} \\ \rho_{j} > \rho_i}} \frac{dist(\mathbf{x}_i^{(z)}, \mathbf{x}_j^{(z)})}{\bar{d}_j^{(z)}},
\end{equation}
where the condition $\rho_{j} > \rho_i$ refers to finding a point $\mathbf{x}_j^{(z)}$ with a higher local density than point $\mathbf{x}_i^{(z)}$. The function $dist(\mathbf{x}_i^{(z)}, \mathbf{x}_j^{(z)})$ represents the Euclidean distance between data points $\mathbf{x}_i^{(z)}$ and $\mathbf{x}_{j}^{(z)}$, while  $\bar{d}_j^{(z)}$ denotes the average distance between all point pairs in subcluster $C_j^{(z)}$. The selection of new seed points is then based on the maximum density gap $\delta_j$ between peaks within the subcluster and its corresponding number of wins $n_j$, using the criterion: 
\begin{equation}
j^*=\arg\max_j \, n_j \cdot \delta_j,
\end{equation}
where $n_j$ is the number of times subcluster $j$ has won, and $\delta_j$ is its maximum inter-density peak distance. Finally, the algorithm terminates when the maximum movement distance of all seed points is less than a threshold $\xi$, indicating convergence:
\begin{equation}
\max_j \left\| \mathbf{m}_j^{\text{old}} - \mathbf{m}_j \right\|_2 < \xi,
\end{equation}
where $\mathbf{m}_j^{\text{old}}$ is the seed point position from the previous iteration, and $\xi$ is a small positive constant that controls convergence precision. Through this algorithm, the data can be divided into multiple micro subclusters on the client.

After completing the local clustering, each client obtains several subclusters. To ensure data privacy, a transmission strategy is designed in which the means and covariances of these subclusters are computed locally. An equivalent amount of data is then randomly generated on the client side based on a multivariate normal distribution, which replaces the original data for transmission to the server. This synthetic data enables the server to perform the subsequent clustering phase without accessing the raw data.

This approach protects data privacy without transmitting the original data while retaining a rough approximation of the data distribution. Additionally, the radius and standard deviation of each subcluster can be computed in advance for subsequent use. The relevant parameter calculations are as follows: the radius of a subcluster is defined as the average distance from all points within the subcluster to the centroid. The calculation formula is 
\begin{equation}\label{eq:r}
r_g^{(z)} = \frac{1}{n_g^{(z)}} \sum_{\mathbf{x}_i^{(z)} \in C_g^{(z)}} dist(\mathbf{x}_i^{(z)}, \mathbf{c}_g^{(z)}), 
\end{equation}
where $n_g^{(z)} = |C_g^{(z)}|$ denote the number of samples in the $g$-th subcluster of client $z$, and the total number of local samples satisfies $n^{(z)}= \sum_{g=1}^{k} n_g^{(z)}$. The centroid of a subcluster is the mean of all data points within the subcluster, calculated as follows:
\begin{equation}\label{eq:c}
    \mathbf{c}_g^{(z)} = \frac{1}{n_g^{(z)}} \sum_{i=1}^{n_g^{(z)}} \mathbf{x}_i^{(z)}. 
\end{equation}The standard deviation is the standard deviation of the data points within the subcluster, calculated as follows:
\begin{equation}\label{eq:s}
    \mathbf{s}_g^{(z)}=\sqrt{\frac{\sum_{i=1}^{n_g^{(z)}}(\mathbf{x}_i^{(z)}-\overline{\mathbf{x}}^{(z)})^{2}}{n_g^{(z)}-1}},\overline{\mathbf{x}}=\frac{\sum_{j=1}^{n_g^{(z)}}\mathbf{x}_j^{(z)}}{n_g^{(z)}}. 
\end{equation}
If the data is multidimensional, the standard deviation for each dimension needs to be calculated, and these values are combined into a vector. The calculation for subsequent parameters follows similarly. The mean of the subcluster is given by 
\begin{equation}\label{eq:mean}
    \bm{\mu}_g^{(z)} = \frac{1}{n_g^{(z)}} \sum_{i=1}^{n_g^{(z)}} \mathbf{x}_i^{(z)}. 
\end{equation}
The covariance is given by
\begin{equation}\label{eq:cov}
    \Gamma_g^{(z)} = \frac{1}{n_g^{(z)}-1} \sum_{i=1}^{n_g^{(z)}} (\mathbf{x}_i^{(z)}-\bm{\mu}_g^{(z)})(\mathbf{x}_i^{(z)}-\bm{\mu}_g^{(z)})^T.
\end{equation}
Synthetic samples generated from the $g$-th subcluster of client $z$ follow a multivariate normal distribution:
\begin{equation}\label{eq:gen}
    \mathfrak{X}_g^{(z)} \sim \mathcal{N}(\bm{\mu}_g^{(z)}, {\Gamma}_g^{(z)}), 
\end{equation}
where $\mathfrak{X}_g^{(z)}$ denotes a random vector representing such a sample. After completing the calculations for $R = \{r_1^{(z)}, r_2^{(z)}, \ldots, r_g^{(z)}\}$, $N = \{n_1^{(z)}, n_2^{(z)}, \ldots, n_g^{(z)}\}$, and $S = \{s_1^{(z)}, s_2^{(z)}, \ldots, s_g^{(z)}\}$, and generating the substitute data using the multivariate normal distribution, the data will be uploaded to the server. The summary of ``Client-Side Automated Micro-Partitioning'' is presented in Algorithm \ref{alg1}. 

This section completes the local clustering on the clients and ensures data privacy by transmitting substitute data. The following section introduces the method by which the server determines the number of clusters and merges the subclusters from each client.

\begin{algorithm}[t]
\caption{\enskip Client-Side Automated Micro-Partitioning}\label{alg1}
\begin{algorithmic}[1]
\STATE \textbf{Input: }Dataset $\textbf{X}^{(z)}$ from client $z \in \{1,2, \ldots, Z\}$. 
\STATE On client $z$, we execute SNP with $\textbf{X}^{(z)}$ to obtain subclusters $\{C_1^{(z)}, C_2^{(z)}, \ldots, C_k^{(z)}\}$; 
\STATE Calculate the $R$, $N$ and $S$ corresponding to subclusters by Eqs. (\ref{eq:r}), (\ref{eq:s}); 
\STATE Calculate the $\bm\mu_g^{(z)}$, $\Gamma_g^{(z)}$ by Eqs. (\ref{eq:mean}), (\ref{eq:cov}), and generate substitute data by Eq. (\ref{eq:gen}); 
\STATE Upload $R$, $N$, $S$ and the substitute data to server;
\STATE \textbf{Output: }Subclusters $\{C_1^{(z)}, C_2^{(z)}, \ldots, C_k^{(z)}\}$. 
\end{algorithmic}
\end{algorithm}

\subsection{Server-Side Hierarchical Merging}
On the server side, the goal is to obtain a suitable number of global clusters $k^*$. However, the subclusters uploaded from each client are generated locally and do not contain direct information about $k^*$. Therefore, the server first collects all the uploaded subclusters and calculates their total number, denoted as $k_0$. These subclusters serve as the basis for initializing the global centroids $\{\mathbf{c}_1, \mathbf{c}_2, \ldots, \mathbf{c}_{k_0}\}$ using the centers of these subclusters. Based on this information, an adaptive method, referred to as “Selection of Number of Clusters (SNC)”, is first introduced to determine the appropriate number of clusters. Once $k^*$ is estimated, a merging-based clustering approach is applied to perform global clustering. If traditional partition-based clustering methods are used to process the received data, the “uniform effect” may occur when the data is imbalanced. To avoid this issue, the merging strategy iteratively combines subclusters—obtained from local clustering on each client—based on their similarities until only $k^*$ clusters remain.

Before introducing the SNC algorithm, we first provide some relevant definitions.

\textbf{Definition 1.} Loose Natural Neighbors (LNN): If two points are mutual $b$-nearest neighbors, they are considered to be natural neighbors \cite{Zhu2016NN}, which can be expressed as
\begin{equation}\label{LNN}
    \mathbf{x}_i \in \text{LNN}(\mathbf{x}_j) \iff (\mathbf{x}_i \in \text{NN}_b(\mathbf{x}_j)) \cap (\mathbf{x}_j \in \text{NN}_b(\mathbf{x}_i)), 
\end{equation}
   where $\text{NN}_b(\mathbf{x}_i)$ denotes the set of $b$-nearest neighbors of point $\mathbf{x}_i$, and $\text{LNN}(\mathbf{x}_i)$ denotes the set of natural neighbors of point $\mathbf{x}_i$.

The effectiveness of the natural neighbor algorithm \cite{Zhu2016NN}  has been demonstrated, although limitations arise when it is applied to imbalanced data. Specifically, sample points from clusters with fewer data points may struggle to find correct neighbors due to their lower density. These points are often surrounded by clusters with more data points, causing their neighborhoods to include samples from other clusters. As a result, sample points from clusters with fewer data points are more likely to be incorrectly connected to other clusters in the neighborhood graph. To address this issue, we introduce the concept of ``strict natural neighbors''.

\textbf{Definition 2. }Strict Natural Neighbors (SNN): Building on the concept of loose natural neighbors, if two points are not only mutual $b$-nearest neighbors but also each other's $m$-th nearest neighbor (where $1 \leq m \leq b$), they are referred to as `strict' natural neighbors, which can be expressed as
\begin{eqnarray}\label{SNN}
    \mathbf{x}_i \in \text{SNN}(\mathbf{x}_j) \Leftrightarrow \mathbf{x}_i \in \text{LNN}_b(\mathbf{x}_j) \cap \mathbf{x}_i = (\text{NN}_b^{(m)}(\mathbf{x}_j)) \cap \mathbf{x}_j = \text{NN}_b^{(m)}(\mathbf{x}_i) (1 \leq m \leq b),
\end{eqnarray}
where $\text{NN}_b^{(m)}(\mathbf{x}_i)$ denotes the $m$-th nearest neighbor of point $\mathbf{x}_i$ within its $b$-nearest neighbors. This definition emphasizes the hierarchical and tight relationship between neighbors.

``Strict natural neighbors'' adds a constraint that two points must be each other's $m$-th nearest neighbors ($1 \leq m \leq b$), prioritizing the connection of samples with similar densities, as shown in Figure \ref{SNN}. Compared to the traditional natural neighbor definition, this improvement reduces cross-cluster erroneous connections and strengthens the reliability of connections within the less populated clusters, thereby enhancing its performance in handling imbalanced data.
\begin{figure}[t]
\centerline{\includegraphics[width=1\textwidth]{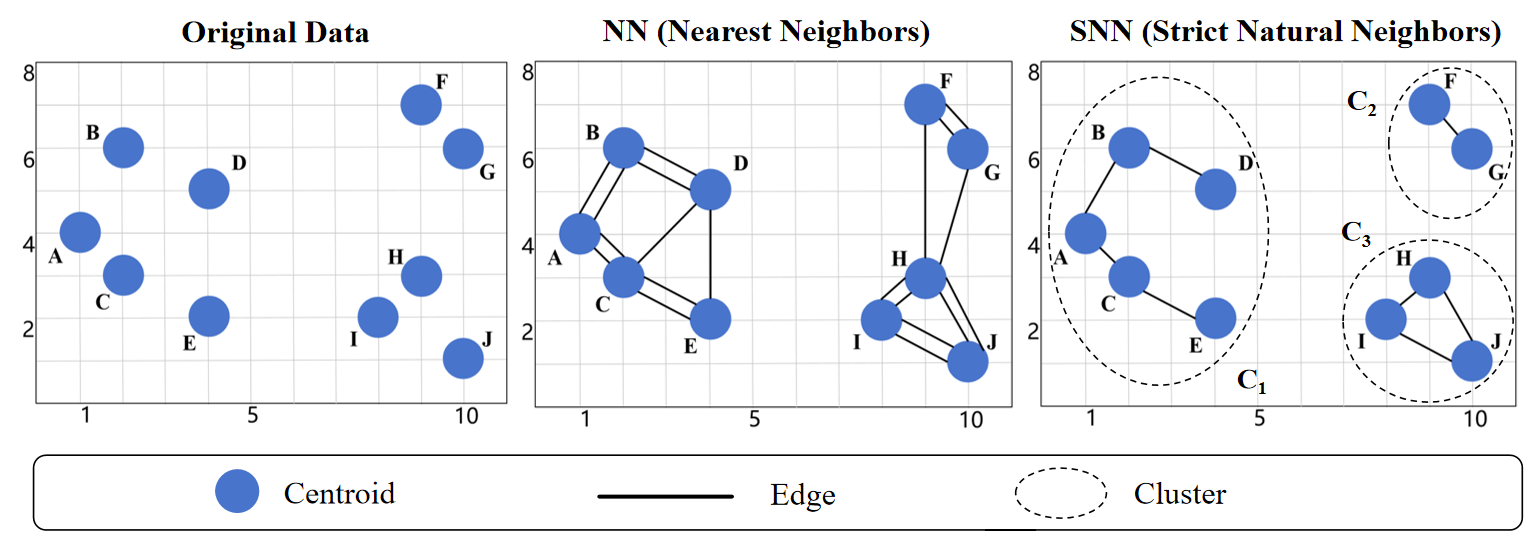}}
\caption{To intuitively illustrate the construction process of SNN, a simplified example is presented using several sample points, and their nearest neighbor (NN) structures are visualized under different neighborhood sizes. In the SNN definition, a pair of points is considered an SNN pair only if they are mutual nearest neighbors, i.e., each lies in the $b$-nearest neighbor list of the other. In the diagram, such mutually connected pairs are denoted with solid lines. For example, under $m$ = 2, the neighbors are as follows:
A → \{B, C\}, B → \{A, D\}, C → \{A, E\}, D → \{A, B\}, E → \{A, C\},F → \{G, H\}, G → \{F, H\}, H → \{I, J\}, I → \{H, J\}, J → \{H, I\}.
Only pairs that appear in both nodes’ neighbor lists qualify as SNN pairs. For instance, since A and B connect each other, the pair (A, B) is an SNN pair, according to Definition 2. In contrast, although D has A as a neighbor, A does not connect D. As a result, (A, D) is not judged as an SNN pair. Accordingly, the identified SNN pairs under $m$ = 2 are: (A, B), (A, C), (B, D), (C, E), (F, G), (H, I), (H, J), (I, J), which form three clusters {$C_{1}, C_{2},C_{3}$}.
\label{SNN}}
\end{figure}

With the above definitions in place, we begin to introduce the SNC algorithm. This algorithm determines the appropriate number of clusters by identifying the neighbor relationships of the centroids corresponding to the subclusters from each client.

First, a coarse-grained recognition method is adopted to determine the primary locations of the centroid distribution. Due to the complex data distribution across clients, where a single cluster might span multiple clients, the local clustering may identify this cluster as multiple subclusters, resulting in increased subcluster density in certain regions on the server, thereby exacerbating data imbalance. To address this issue, existing methods such as the GCS algorithm \cite{Fritzke1994GCS} can be leveraged to represent the subcluster centroids on the server as several main nodes, thus mitigating the degree of data imbalance.

Once the main nodes are obtained, the number of clusters corresponding to these nodes is determined using the proposed neighbor method. First, the distance matrix is computed by computing the pairwise distances between all data points in the dataset, resulting in a distance matrix $\mathbf{D}$ : 
\begin{equation}\label{dm}
\mathbf{D}(i, j) = \| \mathbf{x}_i - \mathbf{x}_j \|_2. 
\end{equation}
Here, $\mathbf{x}_i$ and $\mathbf{x}_j$ represent the feature vectors of the $i$-th and $j$-th points in the dataset. Then, the short-distance threshold $T$ is computed based on the input percentile $t$. The elements in $\mathbf{D}$ are sorted in ascending order as:  
\begin{equation}\label{D^{asc}}D^{asc} = \{ d_{(1)}, d_{(2)}, \dots, d_{(m)} \}, \quad m = \frac{n(n-1)}{2}, 
\end{equation}
where $d_{(1)} \leq d_{(2)} \leq \dots \leq d_{(m)}$ represents the ordered pairwise distances between data points. Next, determine the index $idx$ of $T$ in $D^{asc}$. When the input percentage parameter is $t$ (e.g., $t = 25$ represents $25\%$), then 
\begin{equation}\label{idx}idx = \lceil t \cdot m / 100 \rceil, 
\end{equation}
where $\lceil \cdot \rceil$ denotes the ceiling function. This formula identifies the $idx$-th smallest distance as the short-distance threshold: 
\begin{equation}\label{T}T = d_{(idx)}. 
\end{equation}
Pairs of points with distances smaller than $T$ are referred to as ``short-distance pairs''. To determine an appropriate neighbor size $b$ used in the $b$-nearest neighbors search, the algorithm iteratively evaluates each candidate value $b \in \{{1,2, \ldots, b_{max}}\}$, where $b_{max}$ is a predefined upper bound satisfying $1 < b_{max} < n$. During this process, the algorithm identifies the loose natural neighbors for each point and calculates the ratio $P$ of short-distance pairs among these loose natural neighbor pairs: 
\begin{equation}\label{p}
    P_b = \frac{N_{\text{short}}}{N_{\text{total}}}, 
\end{equation}
where $N_{\text{short}}$ is the number of short-distance pairs and $N_{\text{total}}$ is the number of loose natural neighbor pairs. 

When the ratio $P_b$ begins to decline, it indicates the presence of more distant natural neighbors, prompting the algorithm to stop the iteration and select the previous $b$ value as the optimal value $b_{optimal}$. Based on $b_{optimal}$, the algorithm identifies the strict natural neighbors for each data point and calculates the average distance of all strict natural neighbor pairs, denoted as $D_s$: 
\begin{equation}\label{ne}
    D_s = \frac{1}{N} \sum_{j=1}^{N} \mathbf{D}(p_j, q_j), 
\end{equation}
where $N$ is the number of strict natural neighbor pairs, and $p_j$ and $q_j$ are the indices of the strict natural neighbor pairs. Subsequently, based on the neighbor eigenvalue, we generate the adjacency matrix, denoted as $\mathbf{A}$: 
\begin{equation}\label{am}
    \mathbf{A}(i, j) = 
\begin{cases} 
1 & \text{if } \mathbf{D}(i, j) \leq D_s \\
0 & \text{otherwise}
\end{cases}. 
\end{equation}
Finally, use the adjacency matrix $\mathbf{A}$ to create a graph $G$, and calculate the number of connected components in the graph, which will be the number of clusters $k^*$. The summary of “ SNC” is presented in Algorithm \ref{alg2}.

\begin{algorithm}[t]
\caption{\enskip SNC: Selection of Number of Clusters}\label{alg2}
\begin{algorithmic}[1]
\STATE \textbf{Input: }A set of feature vectors $\mathbf{X} = \{\mathbf{x}_1, \mathbf{x}_2, \ldots, \mathbf{x}_n\}$, $t$: the short-distance threshold percentile.
\STATE Run GCS with $\mathbf{X}$ and obtain the corresponding nodes; 
\STATE Calculate $\mathbf{D}$ based on $nodes$ by Eq. (\ref{dm}); 
\STATE Calculate the short-distance threshold $T$ based on $t$ and $D^{asc}$;
\STATE Initialize $b_{max}$; 
\FOR {$b$ $\leftarrow$ 1 to $b_{max}$}
\STATE Find the loose natural neighbors of each point with $b$;
\STATE Calculate the ratio of short-distance pairs: $P_{b} = \frac{N_{\text{short}}}{N_{\text{total}}};$
\IF {$P_{b-1} > P_{b}$}
\STATE $b_{optimal} = b-1$; 
\STATE break;
\ENDIF
\ENDFOR
\STATE Find the strict natural neighbors of each point with $b_{optimal}$;
\STATE Calculate the neighbor average distance: $D_s = \frac{1}{N} \sum_{j=1}^{N} \mathbf{D}(p_j, q_j)$; 
\STATE Generate the adjacency matrix based on $\mathbf{A}$: \\ $\mathbf{A}(i, j) = 
\begin{cases} 
1 & \text{if } \mathbf{D}(i, j) \leq D_s \\
0 & \text{otherwise}
\end{cases}$; 
\STATE Create the graph $G$ using $\mathbf{A}$, and calculate the number of connected components in $G$, which corresponds to the number of clusters $k^*$; 
\STATE \textbf{Output: }Number of clusters $k^*$. 
\end{algorithmic}
\end{algorithm}

After determining the number of clusters $k^*$, the identified subclusters are merged. To effectively evaluate the similarity between two subclusters, a distance formula is proposed:
\begin{equation}\label{eq:d}
d_{C_i, C_j} = dist(\mathbf{c}_i, \mathbf{c}_j) \cdot o_{C_i,C_j} \cdot sim_{C_i,C_j}^{\beta}. \end{equation}
In this formula, a smaller $d$ value indicates a smaller special distance between the two corresponding subclusters, which in turn suggests greater similarity between them. This is crucial for the subsequent aggregation process. The coefficient $o$ represents the overlap degree between two subclusters. Assuming these clusters are circularly distributed, $r_1$ and $r_2$ represent the radii of the two circles, while $d$ is the distance between their centers. It is important to note that $r_1 + r_2$ is a fixed value, and the ratio of $d$ to $r_1 + r_2$ fluctuates with the degree of overlap between the two circles. When the circles do not overlap, the ratio is greater than 1; conversely, when they overlap, the ratio becomes less than 1, with a smaller ratio indicating a greater degree of overlap. This means the value of $o$ in the formula decreases, suggesting that the two subclusters are more likely to be merged into the same cluster. Therefore, we define the overlap degree $o$ as follows:
\begin{equation}\label{eq:o}
o_{C_i,C_j} = \frac{dist(\mathbf{c}_i, \mathbf{c}_j)}{r_{C_i} + r_{C_j}}.
\end{equation}
While this approach is relatively intuitive, subsequent experimental results demonstrate its effectiveness in practice. Further optimization of this formula can be explored to enhance the overall performance and effectiveness of the algorithm, thereby producing more accurate clustering results. The similarity measure between two subclusters, denoted as $sim$, is calculated as follows:
\begin{equation}\label{eq:sim}
\mathrm{sim}_{C_i, C_j}^\beta = |\mathbf{s}_{C_i} - \mathbf{s}_{C_j}|^\beta, 
\end{equation}
where $\mathbf{s}_{C_i}$ and $\mathbf{s}_{C_j}$ represent the standard deviations of the two subclusters, respectively. The parameter $\beta$ is a modulation factor that controls the influence of the similarity measure in the distance coefficient calculation. The value of $\beta$ should be greater than or equal to 0. As $\beta$ increases, the impact of the similarity measure on the calculation of $d$ also intensifies. However, if $\beta$ is too large, the result for $d$ may primarily reflect the similarity between the two subclusters, neglecting the distance between their centroids and the degree of overlap. Conversely, if $\beta$ is too small, $sim$ approaches 1, meaning the calculation of $d$ will largely depend on the distance and overlap between the cluster centers. Thus, the process typically starts with $\beta = 1$, followed by fine adjustments based on the specific dataset. When two subclusters are merged, some key parameters for the newly formed cluster $C_m$ need to be calculated. The calculation of the size of $C_m$ (denoted as $n_m$) is straightforward and can be obtained using the following formula:
\begin{equation}\label{eq:n}
n_m = n_p + n_q, 
\end{equation}
where \( n_p = |C_p| \) and \( n_q = |C_q| \) denote the number of data points in the two subclusters \( C_p \) and \( C_q \) to be merged, respectively. Since the substitute data are available on the server, they can be used to calculate other important parameters of $C_m$. Specifically, the radius (denoted as $r_m$), centroid (denoted as $c_m$), and standard deviation (denoted as $s_m$) based on Eqs. (\ref{eq:r}), (\ref{eq:c}), and (\ref{eq:s}). 

After defining these formulas, a subcluster merging algorithm is designed. Here is the algorithm's running process on the server. First, the total number of subclusters on the server is counted, denoted as $k_0$, which also represents the current number of remaining clusters on the server. Next, the centroid set is initialized. The algorithm then enters an iterative loop to merge subclusters. In each iteration of the loop, we use Eqs. (\ref{eq:d}), (\ref{eq:o}), (\ref{eq:sim}) to calculate the pairwise distances between all subclusters. Subsequently, the two closest subclusters, $C_p$ and $C_q$, are selected based on formula (\ref{eq:d}) using the following selection method:
\begin{equation}\label{eq:argmin}
    p, q = \underset{1\leq i,j\leq k_0, i\neq j}{\operatorname*{argmin}}\ d_{C_i,C_j}. 
\end{equation}
After that, these two subclusters are merged into a new subcluster:
\begin{equation}\label{eq:Cnew}
    C_m = C_{p} \cup C_{q}. 
\end{equation}
Once the merging is completed, we calculate the radius, centroid, standard deviation, and size of the new cluster using formulas (\ref{eq:r}), (\ref{eq:c}), (\ref{eq:s}) and (\ref{eq:n}), and update the corresponding values to prepare for the next merging operation.

After $k_0 - k^*$ iterations, the server obtains $k^*$ final clusters. Finally, the global clustering assignments are sent back to the corresponding clients to update the local clustering results, thus completing the clustering task. The summary of “Server-Side Hierarchical Merging” is presented in Algorithm \ref{alg3}.

Overall, the section presents a method to adaptively determine the number of clusters on the server and a method for merging subclusters. Through these approaches, the global clustering task for imbalanced data from various clients is accomplished.

\begin{algorithm}[t]
\caption{\enskip Server-Side Hierarchical Merging}\label{alg3}
\begin{algorithmic}[1]
\STATE \textbf{Input: }subclusters, $t$, $\beta$.
\STATE Calculate the current total number of subclusters $k_0$ on the server and initialize centroids $\{\mathbf{c}_1, \mathbf{c}_2, \ldots, \mathbf{c}_{k_0}\}$ by using subclusters;
\STATE Run SNC with centroids and $t$ to obtain $k^*$;
\WHILE {$k_0 > k^*$}
\STATE Calculate the pairwise distances between all subclusters by Eq. (\ref{eq:d});
\STATE Select the two subclusters $C_p$ and $C_q$ with the smallest distance by Eq. (\ref{eq:argmin});
\STATE Merge $C_p$ and $C_q$ as a new subcluster by Eq. (\ref{eq:Cnew});
\STATE Calculate $r_m$, $\mathbf{c}_m$, $\mathbf{s}_m$, and $n_g$ by Eqs. (\ref{eq:r}), (\ref{eq:c}), (\ref{eq:s}), and (\ref{eq:n});
\STATE Update the corresponding values of $C_m$, $r_m$, $\mathbf{c}_m$, $\mathbf{s}_m$, and $n_m$;
\STATE $k_0 = k_0 -1$; 
\STATE $C_g = \{C_1, \ldots, C_{k_0}\}$; 
\ENDWHILE
\STATE \textbf{Output: }$C_g = \{C_1, \ldots, C_{k^*}\}$. 
\end{algorithmic}
\end{algorithm}

\subsection{Discussion}
In this section, the proposed method is introduced. It consists of local clustering on the client,  determining the number of clusters, and global clustering on the server. By avoiding the occurrence of the ``uniform effect'', the proposed approach demonstrates superior clustering performance when handling imbalanced data. 

On the client, a competitive learning method is employed to adaptively partition the data into multiple micro clusters of similar sizes. To ensure data privacy, a strategy based on a multivariate Gaussian distribution is adopted to generate substitute data $-$ new random data generated from the original data, which is then uploaded to the server for processing. This method not only ensures the privacy of the original data but also preserves the general distribution characteristics of the data. In determining the number of clusters, the proposed SNC algorithm combines both loose and strict natural neighbor criteria. A loose criterion is used to select the appropriate $b$-nearest neighbor parameter to enhance the algorithm's tolerance for complex data handling and improve computational efficiency, while a strict criterion is employed when calculating the final neighbor eigenvalue to ensure clustering accuracy. This strategy effectively identifies the cluster structure in the dataset and provides a reliable number of clusters $k^*$ for the subsequent merging process. Finally, on the server, subclusters are merged based on their similarity to complete the global clustering.

\begin{remark}
Privacy Protection: The transmission strategy randomly generates an equivalent amount of data using a multivariate normal distribution to replace the original data to be uploaded to the server, allowing the server to conduct the next phase of clustering and effectively mitigates privacy risks. This conforms with the basic privacy protection requirements of federated learning, i.e., the raw samples and their identities should not be leaked. However, it is still possible that the uploaded means and covariances of samples can be utilized to recover the original cluster distribution, especially when subclusters are very small. Therefore, in scenarios with stricter privacy protection requirements, existing privacy-enhancing techniques such as homomorphic encryption \cite{acar2018survey} and differential privacy \cite{wei2020federated,li2023differentially} can be integrated to perturb the uploaded statistics while preserving the effectiveness of the hierarchical merging on the server. 
\end{remark}

\section{Experiments}\label{sec4}
All experiments were performed on a workstation equipped with a 12th Gen Intel® Core™ i7-12700 CPU running at 2.10 GHz and 32 GB RAM. 

In this section, the following experiments were conducted: 1) A comparative study of the proposed method against various baseline methods on different datasets to comprehensively evaluate the clustering performance of the approach; 2) Ablation studies from both dataset and method perspectives to validate the effectiveness of the method in handling imbalanced data and adaptively determining the number of clusters; 3) A parameter sensitivity analysis for the two main parameters in the algorithm; 4) An evaluation of the runtime of each algorithm. A detailed analysis of these experiments is presented below.

\subsection{Datasets and Comparison Methods}
To comprehensively evaluate the clustering performance of our method, Fed-$k^*$-HC was compared with five other federated clustering algorithms across multiple datasets. The six synthetic datasets include $ids2$ \cite{liang2012k} and $gaussian$, both generated using a mixture of bivariate Gaussian density functions, similar to the generation method in SMCL [39]. Two modified versions, $ids2\_22$ and $ids2\_2k22$, are derived from the $ids2$ dataset. The remaining two datasets, $ids2\_non\_iid$ and $gaussian\_non\_iid$, are constructed based on $ids2$ and $gaussian$, respectively, but follow a non-independent and identically distributed (non-IID) setting to simulate data heterogeneity across clients. Additionally, five real-world datasets from the UCI Machine Learning Repository \cite{uci} were selected for evaluation. The UCI repository is a widely used and publicly available collection of machine learning datasets. In all experiments, the features of all datasets were standardized, and the values were mapped to the [0, 1] range. For detailed information about these datasets, please refer to Table \ref{dataset}.

\begin{table}[!t]
\caption{Information of the datasets involved in our experiments, including five real-world datasets and six synthetic datasets, all of which are imbalanced.}
\centering
\begin{tabular}{c|ccccc}
\toprule
 No.&Dataset    & $d$ & $n$  & $K$ & Cluster statistics                 \\
\midrule
 1&ids2       & 2       & 3200 &  5  &$\{2000, 400, 400, 200, 200\}$ \\
 2&gaussian   & 2       & 2000 &  4  &$\{61, 1212, 606, 121\}$ \\
 3&ids2\_22   & 2       & 400  &  2  &$\{200, 200\}$ \\
 4&ids2\_2k22 & 2       & 2400 &  3  &$\{2000, 200, 200\}$ \\
 5&ids2\_non\_iid &2&3200& 5  &$\{2000, 400, 400, 200, 200\}$ \\
 6&gaussian\_non\_iid & 2& 2000& 4  &$\{61, 1212, 606, 121\}$ \\
 7& pageblock  & 10      & 5357 & 3  &$\{4913, 329, 115\}$ \\ 
 8& yeast      & 6       & 1394 & 6  &$\{463, 429, 244, 163, 51, 44\}$ \\
 9& abalone    & 7       & 1262 & 4  &$\{115, 391, 689, 67\}$ \\
 10& breast     & 9       & 683  & 2  &$\{444, 239\}$ \\
 11& digits& 16& 5666 & 10 &$\{1144, 1143, 798, 608, 552, $ \\
 & & & & &$\ \ 349, 341, 339, 252, 140\}$ \\ \bottomrule
\end{tabular}
\label{dataset}
\end{table}

In our experiments, five comparison methods were used. First, two state-of-the-art (SOTA) methods were selected: KFed (ICML 2021) \cite{dennis2021heterogeneity}, a federated clustering method based on $k$-means; MUFC (ICLR 2023) \cite{pan2023machine}, a method based on $k$-means++; F3KM (PACMMOD 2023) \cite{ Zhu2023F3KMFF}, a fairness-aware fuzzy k-means approach designed for imbalanced federated data; and Orchestra (ICML 2022) \cite{ Lubana2022OrchestraUF}, a memory-based global-local clustering framework with Sinkhorn label alignment. An enhanced version of Orchestra, denoted as $Orchestra^*$, is proposed in this work as an improvement over the original method \cite{ Lubana2022OrchestraUF}. It further improves minority class recognition by introducing class-proportional Sinkhorn alignment. Additionally, since there are relatively few existing federated clustering methods, three diverse types of federated clustering methods are derived and modified from the existing clustering approaches, including Affinity Propagation \cite{frey2007clustering}, Mean Shift \cite{carreira2015review}, and the $b$-nearest neighbors-based DPC algorithm \cite{long2022clustering}. These three counterparts are included for more comprehensive comparison, as they do not assume spherical clusters or require a predefined number of clusters, instead relying on similarity propagation, density estimation, or local density peaks, respectively. These three counterparts follow the conventional federated clustering settings, i.e., with a predefined cluster number $k$. Their adopted original clustering methods are first implemented for local clustering on the clients, and then the relevant parameters are uploaded to the server for global clustering using $k$-means. These three methods are named as Fed-AP, Fed-MS, and Fed-KDPC, respectively.

\begin{table}[!t]
\caption{Clustering performance comparison on the seven datasets evaluated by the five validity metrics. The best and second-best results are indicated by boldface and underline, respectively. The ``Avg. Rank'' row represents the average rank of the methods across different datasets.}
\centering
\resizebox{\textwidth}{!}
{
\begin{tabular}{cclccccccccc}
\toprule
\multicolumn{2}{c}{Dataset}&                 Measurement& Fed-$k^*$-HC (ours)            & KFed          &MUFC &F3KM &Orchestra &$Orchestra^*$& Fed-AP        & Fed-MS           & Fed-KDPC         \\ \midrule
\multicolumn{2}{c}{\multirow{5}{*}{pageblock}}& F-measure              & \textbf{0.9040$\pm$0.0177 }& 0.8514$\pm$0.0724&0.8579$\pm$0.0007 &0.4630$\pm$0.0000&0.2857$\pm$0.0132 &0.5602$\pm$0.0289& \underline{ 0.8877$\pm$0.0000}& 0.8869$\pm$0.0000& 0.8219$\pm$0.0000\\
\multicolumn{2}{l}{}                                     & Accuracy             & 0.9610$\pm$0.0430& 0.9252$\pm$0.1366&0.9311$\pm$0.0012 &0.8199$\pm$0.0000 &0.4065$\pm$0.0199 &0.8449$\pm$0.0200& \underline{ 0.9985$\pm$0.0000}& \textbf{0.9989$\pm$0.0000}& 0.8391$\pm$0.0000\\
\multicolumn{2}{l}{}                                     & NMI             & \underline{0.1543$\pm$0.0630}& 0.0390$\pm$0.0356&0.0206$\pm$0.0004 &0.1170$\pm$0.0000&0.0493$\pm$0.0086 &\textbf{0.2019$\pm$0.0438}& 0.0891$\pm$0.0000& 0.0744$\pm$0.0000& 0.0202$\pm$0.0000\\
\multicolumn{2}{l}{}                                     & ARI             & \underline{0.2933$\pm$0.1154}& 0.0260$\pm$0.0161&0.0175$\pm$0.0013 &0.1133$\pm$0.0000&0.0105$\pm$0.0018 &\textbf{0.3533$\pm$0.0566}& 0.0312$\pm$0.0000& 0.0234$\pm$0.0000& 0.0502$\pm$0.0000\\
\multicolumn{2}{l}{}                                     & DCV             & 0.1130$\pm$0.0174& 0.1793$\pm$0.1661&\textbf{0.0384$\pm$0.0029} &1.1189$\pm$0.0000 &\underline{0.0510$\pm$0.0316} &1.0038$\pm$0.0431& 0.2103$\pm$0.0000& 0.2112$\pm$0.0000& 0.1837$\pm$0.0000\\ \midrule
\multicolumn{2}{l}{\multirow{5}{*}{yeast}}               & F-measure              & \textbf{0.5791$\pm$0.0316}& 0.4826$\pm$0.0401&0.4994$\pm$0.0027 &0.4478$\pm$0.0000 &0.3582$\pm$0.0152 &0.4348$\pm$0.0225& 0.4657$\pm$0.0315& 0.4125$\pm$0.0000& \underline{ 0.5229$\pm$0.0020}\\
\multicolumn{2}{l}{}                                     & Accuracy             & \underline{ 0.7396$\pm$0.0337}& 0.5882$\pm$0.0846&0.5033$\pm$0.0022 &0.5043$\pm$0.0000 &0.3795$\pm$0.0129 &0.4634$\pm$0.0237& 0.6819$\pm$0.0696& \textbf{0.9620$\pm$0.0000}& 0.5911$\pm$0.0040\\
\multicolumn{2}{l}{}                                     & NMI             & \textbf{0.3281$\pm$0.0304}& 0.2191$\pm$0.0306&0.2640$\pm$0.0037 &\underline{ 0.3016$\pm$0.0000} &
 0.1524$\pm$0.0294 &0.2127$\pm$0.0342& 0.2102$\pm$0.0286& 0.1046$\pm$0.0000& 0.2535$\pm$0.0022\\
\multicolumn{2}{l}{}                                     & ARI             & \textbf{0.2359$\pm$0.0384}& 0.1459$\pm$0.0368&0.1770$\pm$0.0278 &\underline{ 0.2232$\pm$0.0000} &0.0856$\pm$0.0144 &0.1307$\pm$0.0141& 0.1166$\pm$0.0277& 0.0239$\pm$0.0000& 0.1918$\pm$0.0019\\
\multicolumn{2}{l}{}                                     & DCV             & 0.5380$\pm$0.0336& 0.2933$\pm$0.1525&0.2740$\pm$0.0078&0.6317$\pm$0.0000 &\underline{ 0.0571$\pm$0.0144} &0.6160$\pm$0.0247& 0.5242$\pm$0.0810& 1.5558$\pm$0.0000& \textbf{0.0203$\pm$0.0138}\\ \midrule
\multicolumn{2}{l}{\multirow{5}{*}{abalone}}             & F-measure              & \textbf{0.6855$\pm$0.0193}& 0.5334$\pm$0.0275&0.5233$\pm$0.0008 &0.4571$\pm$0.0000 &0.3801$\pm$0.0317 &0.5094$\pm$0.0087& 0.5461$\pm$0.0062& 0.5155$\pm$0.0000& \underline{ 0.5632$\pm$0.0112}\\
\multicolumn{2}{l}{}                                     & Accuracy             & \textbf{0.7274$\pm$0.0444}& 0.5171$\pm$0.0283&0.4995$\pm$0.0016 &0.5008$\pm$0.0000 &0.4155$\pm$0.0330 &0.5900$\pm$0.0067& 0.5103$\pm$0.0083& \underline{0.7124$\pm$0.0000}& 0.5380$\pm$0.0127\\
\multicolumn{2}{l}{}                                     & NMI             & \textbf{0.3019$\pm$0.0099}& 0.2553$\pm$0.0318&0.2548$\pm$0.0006 &0.2638$\pm$0.0000 &0.2289$\pm$0.0099 &\underline{0.2794$\pm$0.0139}& 0.2703$\pm$0.0003& 0.1525$\pm$0.0000& 0.2735$\pm$0.0046\\
\multicolumn{2}{l}{}                                     & ARI             & \textbf{0.2868$\pm$0.0315}& 0.1682$\pm$0.0305&0.1527$\pm$0.0005 &0.1617$\pm$0.0000 &0.1387$\pm$0.0060 &0.1755$\pm$0.0063& 0.1849$\pm$0.0035& 0.0214$\pm$0.0000& \underline{ 0.1922$\pm$0.0047}\\
\multicolumn{2}{l}{}                                     & DCV             & \textbf{0.0251$\pm$0.0001}& 0.6048$\pm$0.1290&0.5828$\pm$0.0008 &0.2462$\pm$0.0000&\underline{ 0.0313$\pm$0.0253} &0.7303$\pm$0.0380& 0.3272$\pm$0.0070& 0.3446$\pm$0.0000& 0.6363$\pm$0.0509\\ \midrule
\multicolumn{2}{l}{\multirow{5}{*}{breast}}              & F-measure              &0.9517$\pm$0.0069& 0.9580$\pm$0.0018&\underline{0.9612$\pm$0.0007} &0.9496$\pm$0.0000 &0.8707$\pm$0.0096 &0.9175$\pm$0.0154& 0.7957$\pm$0.0170& 0.6888$\pm$0.0000& \textbf{ 0.9663$\pm$0.0000}\\
\multicolumn{2}{l}{}                                     & Accuracy             & 0.9356$\pm$0.0129& 0.9582$\pm$0.0018&\underline{0.9613$\pm$0.0007} &0.9546$\pm$0.0000 &0.8747$\pm$0.0096 &0.9215$\pm$0.0172& 0.8386$\pm$0.0048& 0.9444$\pm$0.0000& \textbf{ 0.9663$\pm$0.0000}\\
\multicolumn{2}{l}{}                                     & NMI             & 0.7243$\pm$0.0262& 0.7368$\pm$0.0081&\underline{0.7519$\pm$0.0033} &0.7206$\pm$0.0000 &0.5576$\pm$0.0199 &0.6671$\pm$0.0395& 0.3867$\pm$0.0287& 0.1437$\pm$0.0000& \textbf{ 0.7749$\pm$0.0000}\\
\multicolumn{2}{l}{}                                     & ARI             & 0.8384$\pm$0.0149& 0.8426$\pm$0.0065&\underline{0.8498$\pm$0.0027} &0.8247$\pm$0.0000 &0.5611$\pm$0.0290 &0.7104$\pm$0.0509& 0.3769$\pm$0.0379& 0.1002$\pm$0.0000& \textbf{ 0.8686$\pm$0.0000}\\
\multicolumn{2}{l}{}                                     & DCV             & 0.9640$\pm$0.2366&  0.0352$\pm$0.0050&\underline{0.0348$\pm$0.0020} &0.3324$\pm$0.0000 &0.0495$\pm$0.0193 &0.1438$\pm$0.0311& 0.5176$\pm$0.0377& 0.8324$\pm$0.0000& \textbf{0.0041$\pm$0.0000}\\ \midrule
\multicolumn{2}{l}{\multirow{5}{*}{digits}}              & F-measure              & \textbf{0.7078$\pm$0.0138}& 0.6227$\pm$0.0260&0.6920$\pm$0.0009 &0.5961$\pm$0.0000 &0.4623$\pm$0.0356 &0.5225$\pm$0.0092& \underline{0.7054$\pm$0.0198}& 0.6819$\pm$0.0036& 0.6845$\pm$0.0084\\
\multicolumn{2}{l}{}                                     & Accuracy             & \textbf{0.7631$\pm$0.0395}& 0.6221$\pm$0.0243&0.6670$\pm$0.0011 &0.6445$\pm$0.0000 &0.4993$\pm$0.0335 &0.6302$\pm$0.0122& \underline{0.7540$\pm$0.0233}& 0.6614$\pm$0.0012& 0.6691$\pm$0.0097\\
\multicolumn{2}{l}{}                                     & NMI             & \textbf{0.7010$\pm$0.0117}& 0.5930$\pm$0.0289&0.6727$\pm$0.0007 &0.6626$\pm$0.0000&0.4199$\pm$0.0485 &0.5528$\pm$0.0254& \underline{0.6980$\pm$0.0096}& 0.6665$\pm$0.0027& 0.6787$\pm$0.0090\\
\multicolumn{2}{l}{}                                     & ARI             & \textbf{0.5686$\pm$0.0163}& 0.4582$\pm$0.0329&0.5418$\pm$0.0011 &0.5350$\pm$0.0000&0.2900$\pm$0.0423 &0.4527$\pm$0.0244& \underline{0.5657$\pm$0.0161}& 0.5303$\pm$0.0093& 0.5369$\pm$0.0079\\
\multicolumn{2}{l}{}                                     & DCV             & \underline{0.1299$\pm$0.0356}&  0.1486$\pm$0.0770&0.2886$\pm$0.0018 &0.3347$\pm$0.0000&\textbf{0.0653$\pm$0.0206} &0.6087$\pm$0.0172& 0.1866$\pm$0.0206& 0.2513$\pm$0.0506& 0.1841$\pm$0.0376\\ \midrule
\multicolumn{2}{l}{\multirow{5}{*}{\makecell[l]{gaussian\\(synthetic)}}} & F-measure              & \textbf{0.9863$\pm$0.0000}& 0.8468$\pm$0.0496&0.8661$\pm$0.0025 &0.7318$\pm$0.0000&0.5180$\pm$0.0170 &0.8436$\pm$0.0184& 0.8212$\pm$0.0401& 0.7086$\pm$0.0000& \underline{ 0.9181$\pm$0.0162}\\
\multicolumn{2}{l}{}                                     & Accuracy             & \textbf{0.9860$\pm$0.0000}& 0.8167$\pm$0.0680&0.8250$\pm$0.0029 &0.8260$\pm$0.0000&0.5867$\pm$0.0185 &\underline{0.9388$\pm$0.0062}& 0.7775$\pm$0.0481& 0.6350$\pm$0.0000& 0.9013$\pm$0.0206\\
\multicolumn{2}{l}{}                                     & NMI             & \textbf{0.9229$\pm$0.0000}& 0.6751$\pm$0.0354&0.7137$\pm$0.0028 &0.7073$\pm$0.0000&0.4277$\pm$0.0699 &\underline{0.7988$\pm$0.0248}& 0.6745$\pm$0.0417& 0.5184$\pm$0.0000& 0.7646$\pm$0.295\\
\multicolumn{2}{l}{}                                     & ARI             & \textbf{0.9682$\pm$0.0000}& 0.6825$\pm$0.1098&0.6479$\pm$0.0051 &0.6492$\pm$0.0000&0.2969$\pm$0.0555 &\underline{ 0.8658$\pm$0.0163}& 0.5710$\pm$0.0737& 0.4682$\pm$0.0000& 0.7761$\pm$0.0440\\
\multicolumn{2}{l}{}                                     & DCV             & \textbf{0.0225$\pm$0.0000}& 0.3149$\pm$0.2253&0.4583$\pm$0.0082 &0.5295$\pm$0.0000&\underline{ 0.0324$\pm$0.0276}&0.8521$\pm$0.0169& 0.5901$\pm$0.1434& 0.5396$\pm$0.0000& 0.2614$\pm$0.0605\\ \midrule
\multicolumn{2}{l}{\multirow{5}{*}{\makecell[l]{ids2\\(synthetic)}}}     & F-measure              & \textbf{0.9935$\pm$0.0000}& \underline{0.7708$\pm$0.0557}&0.7384$\pm$0.0043 &0.5439$\pm$0.0000&0.4967$\pm$0.0330 &0.5324$\pm$0.0386& 0.7586$\pm$0.0149& 0.5950$\pm$0.0000& 0.7558$\pm$0.0063\\
\multicolumn{2}{l}{}                                     & Accuracy             & \textbf{0.9934$\pm$0.0000}& \underline{0.7440$\pm$0.0724}&0.6841$\pm$0.0047 &0.6241$\pm$0.0000&0.5167$\pm$0.0580&0.6734$\pm$0.0366& 0.7068$\pm$0.0179& 0.5544$\pm$0.0000& 0.6973$\pm$0.0073\\
\multicolumn{2}{l}{}                                     & NMI             & \textbf{0.9669$\pm$0.0000}& 0.6488$\pm$0.0412&0.7487$\pm$0.0016 &0.7448$\pm$0.0000&0.4783$\pm$0.0210&0.5260$\pm$0.0530& 0.7694$\pm$0.0011& 0.3145$\pm$0.0000& \underline{ 0.7859$\pm$0.0070}\\
\multicolumn{2}{l}{}                                     & ARI             & \textbf{0.9814$\pm$0.0000}& \underline{0.5550$\pm$0.0914}&0.5261$\pm$0.0013 &0.5286$\pm$0.0000&0.2350$\pm$0.0211 &0.4259$\pm$0.0552& 0.5347$\pm$0.0061& 0.2916$\pm$0.0000& 0.5506$\pm$0.0048\\
\multicolumn{2}{l}{}                                     & DCV             & \textbf{0.0173$\pm$0.0000}& 0.4341$\pm$0.1766&0.7261$\pm$0.0018 &0.4250$\pm$0.0000&\underline{ 0.0548$\pm$0.0222} &0.8814$\pm$0.0551& 0.6998$\pm$0.0159& 0.2495$\pm$0.0000& 0.6954$\pm$0.0056\\ \midrule
 \multicolumn{3}{c}{Avg. Rank}& \textbf{2.0000}& 4.6571&4.6000& 5.6286& 7.2000&5.9429& 4.9714& 6.5714&\underline{3.4286}\\
\bottomrule
\end{tabular}
}
\label{iid test}
\end{table}

\subsection{Evaluation Metrics}
To conduct a comprehensive and intuitive numerical comparison, five evaluation metrics were used to assess the quality of clustering results. These metrics include: 1) F-measure; 2) Accuracy; 3) Normalized Mutual Information (NMI); 4) Adjusted Rand Index (ARI); and 5) DCV. The first four metrics are standard evaluation methods commonly used in clustering analysis \cite{NMI_65fed2d86db441a3919e495dfd0258a6}, while DCV \cite{liang2012k} is specifically designed to evaluate clustering performance on imbalanced data. Although a lower DCV value does not always indicate excellent clustering performance, a higher DCV value typically suggests poorer clustering results. Based on the clustering results of each method on different datasets, the average ranking of each evaluation metric was calculated to provide a comprehensive assessment of the performance of the methods.

\begin{table}[!t]
\caption{Clustering performance comparison on the two Non-IID datasets.}
\centering
\resizebox{\textwidth}{!}{
\begin{tabular}{cclccccccccc}
\toprule
\multicolumn{2}{c}{Dataset}&            Measurement& Fed-$k^*$-HC (ours)            & KFed    &MUFC &F3KM &Orchestra &$Orchestra^*$& Fed-AP  & Fed-MS        & Fed-KDPC \\ \midrule
\multicolumn{2}{l}{\multirow{5}{*}{ids2\_non\_iid}}     & F-measure              & \textbf{0.9984$\pm$0.0000}& 0.9840$\pm$0.0004&0.7413$\pm$0.0034 &0.5435$\pm$0.0000 &0.5058$\pm$0.0092 &0.6034$\pm$0.0551& 0.7177$\pm$0.0373& \underline{0.9871$\pm$0.0000}& 0.7696$\pm$0.0100\\
\multicolumn{2}{l}{}                          & Accuracy             & \textbf{0.9984$\pm$0.0000}& 0.9836$\pm$0.0004&0.6873$\pm$0.0038 &0.6231$\pm$0.0000 &0.5217$\pm$0.0089 &0.7100$\pm$0.0447& 0.6806$\pm$0.0393& \underline{0.9869$\pm$0.0000}& 0.7163$\pm$0.0128\\
\multicolumn{2}{l}{}                          & NMI             & \textbf{0.9917$\pm$0.0000}& 0.9411$\pm$0.0011&0.7486$\pm$0.0012 &0.7440$\pm$0.0000 &0.4664$\pm$0.0388 &0.4339$\pm$0.0357& 0.6799$\pm$0.0301& \underline{0.9496$\pm$0.0000}& 0.7770$\pm$0.0077\\
\multicolumn{2}{l}{}                          & ARI             & \textbf{0.9952$\pm$0.0000}& 0.9534$\pm$0.0010&0.5270$\pm$0.0011 &0.5279$\pm$0.0000 &0.2124$\pm$0.0266 &0.4809$\pm$0.0089& 0.4571$\pm$0.0336& \underline{0.9624$\pm$0.0000}& 0.5467$\pm$0.0024\\
\multicolumn{2}{l}{}                          & DCV             & \textbf{0.0041$\pm$0.0000}& 0.0461$\pm$0.0009&0.7246$\pm$0.0015 &0.4242$\pm$0.0000 &0.0421$\pm$0.0132 &0.7376$\pm$0.0543& 0.7686$\pm$0.0802& \underline{0.0369$\pm$0.0000}& 0.6942$\pm$0.0024\\ \midrule
\multicolumn{2}{l}{\multirow{5}{*}{gaussian\_non\_iid}} & F-measure              & \textbf{0.9960$\pm$0.0000}& 0.9873$\pm$0.0005&0.8663$\pm$0.0021 &0.7318$\pm$0.0000 &0.4616$\pm$0.0570 &0.7218$\pm$0.0377& 0.8923$\pm$0.0267& \underline{0.9921$\pm$0.0000}& 0.8352$\pm$0.0213\\
\multicolumn{2}{l}{}                          & Accuracy             & \textbf{0.9960$\pm$0.0000}& 0.9870$\pm$0.0005&0.8252$\pm$0.0029 &0.8260$\pm$0.0000 &0.5450$\pm$0.0455 &0.8143$\pm$0.0232& 0.8689$\pm$0.0373& \underline{0.9920$\pm$0.0000}& 0.7859$\pm$0.0273\\
\multicolumn{2}{l}{}                          & NMI             & \textbf{0.9761$\pm$0.0000}& 0.9410$\pm$0.0016&0.7135$\pm$0.0014 &0.7073$\pm$0.0000 &0.4601$\pm$0.0333 &0.4502$\pm$0.0503& 0.7081$\pm$0.0305& \underline{0.9584$\pm$0.0000}& 0.7172$\pm$0.0100\\
\multicolumn{2}{l}{}                          & ARI             & \textbf{0.9914$\pm$0.0000}& 0.9742$\pm$0.0007&0.6480$\pm$0.0041 &0.6492$\pm$0.0000 &0.3103$\pm$0.0243 &0.5420$\pm$0.0466& 0.7434$\pm$0.0583& \underline{0.9835$\pm$0.0000}& 0.6022$\pm$0.0273\\
\multicolumn{2}{l}{}                          & DCV             & \textbf{0.0060$\pm$0.0000}& 0.0234$\pm$0.0006&0.4580$\pm$0.0068 &0.5295$\pm$0.0000 &0.0417$\pm$0.0127 &0.7566$\pm$0.0473& 0.2821$\pm$0.0893& \underline{0.0141$\pm$0.0000}& 0.5478$\pm$0.0501\\ \midrule
 \multicolumn{3}{c}{Avg. Rank}& \textbf{1.0000}& 3.1000&5.8000& 6.2000& 7.7000&7.7000& 6.0000& \underline{2.0000}&5.5000\\
 \bottomrule
\end{tabular}}
\label{noniid}
\end{table}

\subsection{Performance Evaluation}
In the federated learning environment, due to the complex and diverse data distributions across different clients, the fairness of the experiments was ensured by first shuffling the original order of the data and evenly distributing it among the clients for federated clustering. This ensured that the data followed an independent and identically distributed (IID) assumption, meaning that the samples were randomly shuffled and evenly partitioned across clients, so that each client received data drawn from the same underlying distribution. The experimental results are presented in Table \ref{iid test}, with the best results highlighted in bold and the second-best results underlined. Experiments were conducted on five real-world datasets and two synthetic datasets. According to the experimental results, our method outperforms most of the evaluation metrics, with Fed-KDPC following closely behind. In contrast, Fed-MS performed the worst, while the other three methods showed varying performances across different datasets. However, on the \textit{pageblock} dataset, both F3KM and the original Orchestra struggled to identify the minority class, resulting in low F1 scores. Although the overall DCV appeared low, this was mainly due to the dominance of majority clusters, masking the imbalance issue. By incorporating class-proportional Sinkhorn alignment into Orchestra (denoted as $Orchestra^*$), the model's ability to recognize minority clusters improved notably, leading to higher F1 and accuracy. However, this also led to a higher DCV, as more samples were assigned to small clusters. In addition, our method performed poorly on the $breast$ dataset. The reason for this is that the boundaries between clusters are not distinct, and the intra-cluster data distribution are relatively sparse. As a result, the SNC algorithm failed to estimate a cluster number close to the ground truth. However, when the real number of clusters $K$ is manually input, the clustering results are better, indicating that the proposed subcluster merging method is effective. In contrast, on the $digits$ dataset, although SNC did not find the exact clustering number, the obtained value of $k^*$ was close to the true value, and thus the clustering performance was not significantly affected. Overall, our method outperforms the other compared methods in addressing the federated clustering problem for imbalanced data.

\begin{table}[!t]
\caption{Clustering performance comparison on two datasets with balanced (ids2\_22) and imbalanced (ids2\_2k22) clusters.}
\centering
\resizebox{\textwidth}{!}{
\begin{tabular}{cclccccccccc}
\toprule
\multicolumn{2}{c}{Dataset}&                   Measurement& Fed-$k^*$-HC (ours)            & KFed        &MUFC &F3KM &Orchestra  &$Orchestra^*$& Fed-AP           & Fed-MS           & Fed-KDPC         \\ \midrule
\multicolumn{2}{l}{\multirow{5}{*}{\makecell[l]{ids2\_22\\(balanced)}}}     & F-measure              & \textbf{1.0000$\pm$0.0000}& \textbf{1.0000$\pm$0.0000}&\textbf{1.0000$\pm$0.0000} &\textbf{1.0000$\pm$0.0000}  &\underline{0.9690$\pm$0.0185} &\textbf{1.0000$\pm$0.0000}  
& 0.6017$\pm$0.0487& 0.6656$\pm$0.0000& \textbf{1.0000$\pm$0.0000}\\
\multicolumn{2}{l}{}                                        & Accuracy             & \textbf{1.0000$\pm$0.0000}& \textbf{1.0000$\pm$0.0000}&\textbf{1.0000$\pm$0.0000} &\textbf{1.0000$\pm$0.0000}  &0.9690$\pm$0.0185 &\textbf{1.0000$\pm$0.0000}  
& 0.6700$\pm$0.0158& \underline{ 0.9950$\pm$0.0000}& \textbf{1.0000$\pm$0.0000}\\
\multicolumn{2}{l}{}                                        & NMI             & \textbf{1.0000$\pm$0.0000}& \textbf{1.0000$\pm$0.0000}&\textbf{1.0000$\pm$0.0000} &\textbf{1.0000$\pm$0.0000}  &\underline{0.8100$\pm$0.1054} &\textbf{1.0000$\pm$0.0000}  
& 0.0261$\pm$0.0406& 0.0235$\pm$0.0000& \textbf{1.0000$\pm$0.0000}\\
\multicolumn{2}{l}{}                                        & ARI             & \textbf{1.0000$\pm$0.0000}& \textbf{1.0000$\pm$0.0000}&\textbf{1.0000$\pm$0.0000} &\textbf{1.0000$\pm$0.0000}  &\underline{0.8801$\pm$0.0704} &\textbf{1.0000$\pm$0.0000}  
& 0.0331$\pm$0.0552& 0.0001$\pm$0.0000& \textbf{1.0000$\pm$0.0000}\\
\multicolumn{2}{l}{}                                        & DCV             & \textbf{0.0000$\pm$0.0000}& \textbf{0.0000$\pm$0.0000}&\textbf{0.0000$\pm$0.0000} &\textbf{0.0000$\pm$0.0000}  &\underline{0.0220$\pm$0.0380} &\textbf{0.0000$\pm$0.0000}  & 0.3492$\pm$0.2067& 1.4001$\pm$0.0000& \textbf{0.0000$\pm$0.0000}\\ \midrule
\multicolumn{2}{l}{\multirow{5}{*}{\makecell[l]{ids2\_2k22\\(imbalanced)}}} & F-measure              & \textbf{0.9963$\pm$0.0000}& 0.6942$\pm$0.0733&0.6630$\pm$0.0090 &0.7326$\pm$0.0000 &0.4989$\pm$0.0197 &0.5871$\pm$0.0067& 0.6647$\pm$0.0321& \underline{ 0.8963$\pm$0.0000}& 0.7672$\pm$0.0056\\
\multicolumn{2}{l}{}                                        & Accuracy             & \textbf{0.9963$\pm$0.0000}& 0.6414$\pm$0.0820&0.5849$\pm$0.0087 &0.8104$\pm$0.0000 &0.5188$\pm$0.0178 &0.7849$\pm$0.0134& 0.6091$\pm$0.0296& \underline{ 0.9288$\pm$0.0000}& 0.7197$\pm$0.0068\\
\multicolumn{2}{l}{}                                        & NMI             & \textbf{0.9609$\pm$0.0000}& 0.2944$\pm$0.0582&\underline{0.5318$\pm$0.0176}&0.4614$\pm$0.0000 &0.4825$\pm$0.0335 &0.4048$\pm$0.0129& 0.3396$\pm$0.0489&  0.4945$\pm$0.0000& 0.4204$\pm$0.0036\\
\multicolumn{2}{l}{}                                        & ARI             & \textbf{0.9835$\pm$0.0000}& 0.2418$\pm$0.0893&0.3247$\pm$0.0096 &0.4311$\pm$0.0000 &0.2413$\pm$0.0296 &0.3005$\pm$0.0205& 0.1691$\pm$0.0408& \underline{ 0.6309$\pm$0.0000}& 0.2942$\pm$0.0084\\
\multicolumn{2}{l}{}                                        & DCV             & \textbf{0.0097$\pm$0.0000}& 0.7460$\pm$0.2277&0.8950$\pm$0.0146 &0.6704$\pm$0.0000 &0.0594$\pm$0.0180 &0.9576$\pm$0.0288& 0.8348$\pm$0.0729& \underline{ 0.0361$\pm$0.0000}& 0.6474$\pm$0.0107\\ \midrule
 \multicolumn{3}{c}{Avg. Rank}& \textbf{1.0000}& 3.8000&3.4000& \underline{2.5000}& 4.4000&3.8000& 5.4000& 2.8000&2.9000\\ \bottomrule
\end{tabular}}
\label{datasetInfo}
\end{table}

To further evaluate the ability of these methods to identify clusters distributed across different clients, a cluster absence experiment was also conducted. In this experiment, each client only contains data from a subset of clusters. The synthetic datasets ($ids2$ and $gaussian$) were assigned to three clients, with each client only having data points from two clusters (meaning the data is Non-IID), simulating a typical federated clustering scenario. The experimental results are summarized in Table \ref{noniid}. In this experiment, our proposed Fed-$k^*$-HC method performed the best, followed by Fed-MS and KFed. The performance of the other three methods was slightly inferior. The results show that our method can effectively identify clusters distributed across multiple clients and performs better than the SOTA methods KFed and MUFC. Additionally, based on the DCV metric, the value for Fed-$k^*$-HC is significantly lower than that of the other methods, further demonstrating its superior performance in handling imbalanced data.

Overall, based on a series of experiments, it can be concluded that the proposed Fed-$k^*$-HC demonstrates superior clustering performance on imbalanced data compared to existing SOTA methods and other comparison approaches. This result indicates that the method has successfully achieved its intended goal and shows significant advantages in addressing the issue of handling imbalanced data in federated clustering.

\subsection{Ablation Studies}
On one hand, ablation experiments were conducted on the datasets to demonstrate the clustering performance differences of various methods on both balanced and imbalanced data. The experiments used the $ids2\_22$ and $ids2\_2k22$ datasets, with relevant details provided in Table \ref{dataset}. Specifically, $ids2\_22$ is a balanced dataset containing only two clusters, while $ids2\_2k22$ is an imbalanced dataset created by adding a larger cluster to the former. According to the experimental results in Table \ref{datasetInfo}, on the balanced $ids2\_22$ dataset, our method, KFed, MUFC, and Fed-KDPC all performed well. However, when the data became imbalanced, on the imbalanced $ids2\_2k22$ dataset, only our method maintained good clustering performance, surpassing other comparative methods and achieving the best results. This result further validates the reliability of our method for federated clustering on imbalanced data.

On the other hand, an ablation study was conducted on the SNC algorithm, which is used to determine the number of clusters adaptively. As mentioned earlier, SNC is implemented using GCS \cite{Fritzke1994GCS} combined with our designed neighbor method. In the ablation experiment, the neighbor method in SNC was replaced with the natural neighbor algorithm \cite{Zhu2016NN}, and the results are shown in Table \ref{ablation on methods}. From the results, it can be seen that the $k^*$ obtained using the natural neighbor algorithm is only close to the $K$ in most datasets, while the SNC method is able to accurately determine $k^*$, which matches the ground-truth $K$ in most datasets. However, on the $breast$ dataset, the $k^*$ determined by our method deviates from $K$. This is because the boundaries between clusters are not clearly defined in this dataset, and some of the intra-cluster data are dispersed, leading to difficulty in finding an appropriate $k^*$. Overall, SNC's performance meets expectations. It is capable of adaptively determining the number of clusters in most datasets and yields good results.

\begin{table}[!t]
\caption{Result of ablation studies on the SNC algorithm. The table lists the number of clusters obtained through different methods on various datasets, with results matching the real value indicated in bold. The values in the column $K$ represent the ground-truth number of clusters in the corresponding dataset.}
\centering
\begin{tabular}{l|ccc}
\toprule
Dataset&  $K$&GCS + NN& SNC  (ours)\\ \midrule
pageblock
&  \textbf{3}&1
& \textbf{3} \\
yeast
&  \textbf{6}&4
& \textbf{6} \\
abalone&  \textbf{4}&3
& \textbf{4} \\
breast
&  \textbf{2}&3
& 5
 \\
digits
&  \textbf{10}&6
& 9
 \\ 
ids2\_22&  \textbf{2}&3& \textbf{2}\\ 
 ids2\_2k22& \textbf{3}& \textbf{3}& \textbf{3}\\
 ids2& \textbf{5}& 2& \textbf{5}\\
 ids2\_non\_iid&\textbf{5}& \textbf{5}& \textbf{5} \\
 gaussian& \textbf{4}& \textbf{4}& \textbf{4}\\
 gaussian\_non\_iid& \textbf{4}& \textbf{4}& \textbf{4}\\ \bottomrule
\end{tabular}
\label{ablation on methods}
\end{table}

\begin{figure}[!t]
\centerline{\includegraphics[width=0.8\textwidth]{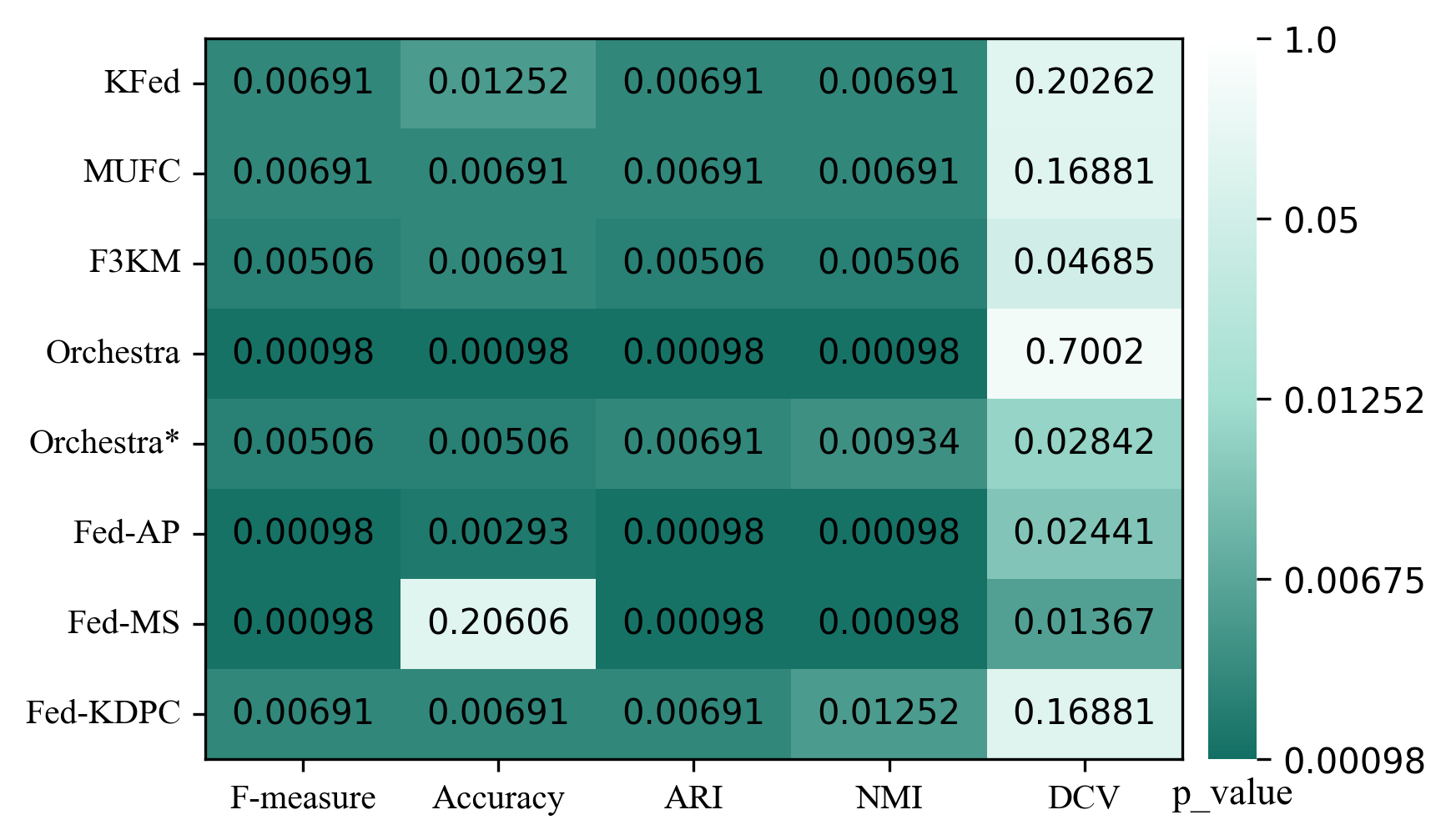}}
\caption{p-values of the Wilcoxon signed rank test in comparing our method against the other methods on five metrics.
\label{p-value}}
\end{figure}

To confirm that improvements over baseline methods are significant and consistent, we conduct significance tests between our method and other methods using the Wilcoxon signed rank test based on the clustering performance reported in Table \ref{iid test}-\ref{datasetInfo}. We show the p-values of the pairwise comparisons in Figure \ref{p-value}, where a darker color represents a smaller p-value. It can be observed that the majority of the comparisons indicate that our method achieves statistically significant superiority at the 99\% confidence level. It is noteworthy that the Wilcoxon signed-rank test is based on the performance rankings of the methods, which leads to relatively coarse-grained p-values.

\subsection{Parameter Sensitivity Studies}
\begin{figure}
\centering
\includegraphics[width = 1\textwidth]{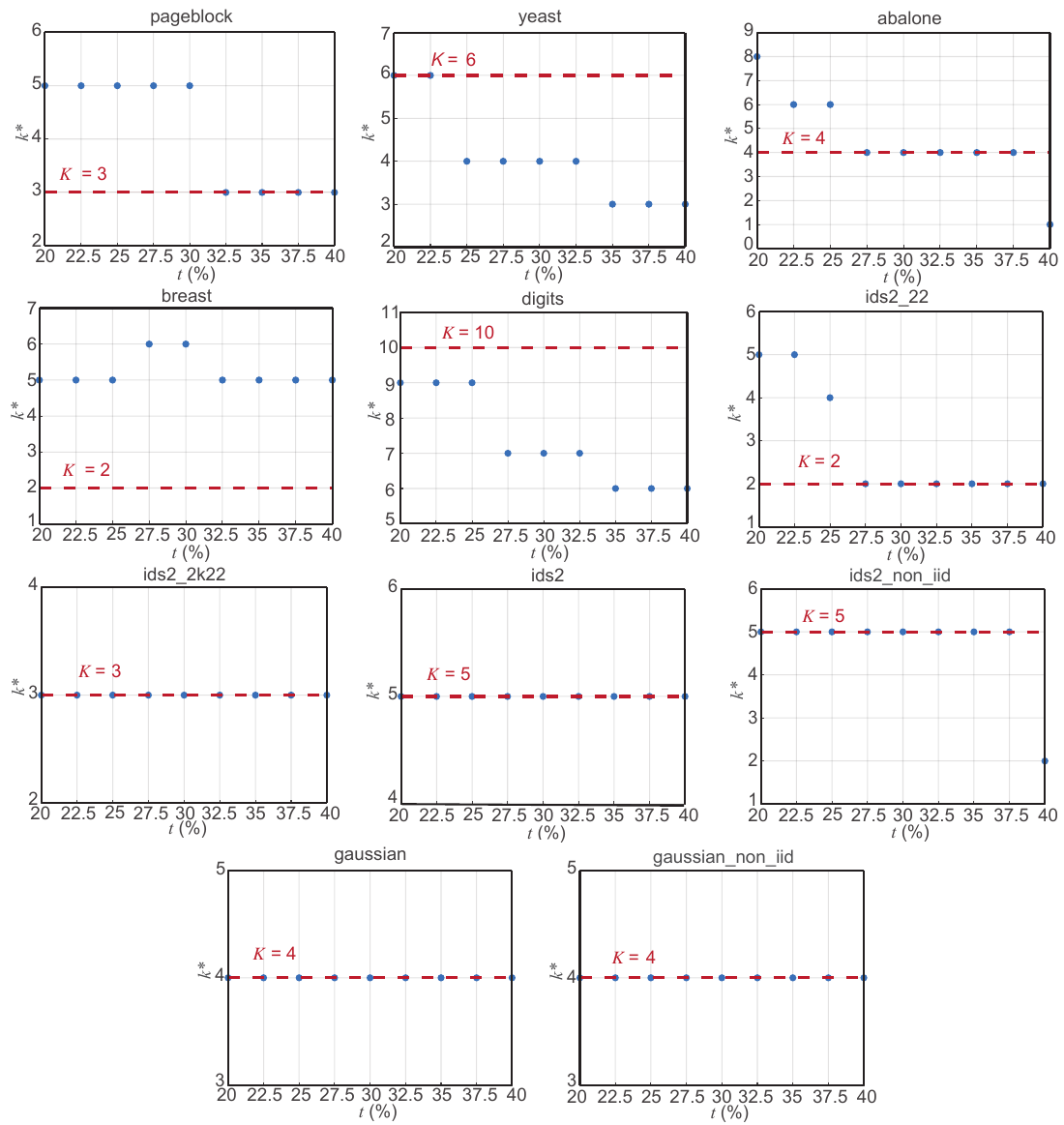}
  \caption{Parameter sensitivity analysis on $t$. The symbol $K$ denotes the ground-truth number of clusters in each dataset. The results are plotted demonstrating how the proposed method can automatically infer the correct cluster number across datasets.} 
  \label{kFt}
\end{figure}

This section evaluates the impact of parameters $t$ and $b$ in SNC, and $\beta$ in the global clustering algorithm on clustering performance. To ensure the reliability of the experimental results, multiple independent runs for each parameter setting are conducted, and the average performance is reported. In the SNC algorithm, sensitivity testing was conducted on the parameter $t$ (the short-distance threshold percentile) to examine the number of clusters $k^*$ obtained under different $t$ values, as shown in Figure \ref{kFt}. Note that the short-distance threshold refers to the value of the distance between sample pairs that falls below the $t$ percentile in the overall distance distribution of the samples. This threshold is used to determine whether an adjacency relationship should be established. In the figure, the x-axis represents the short-distance threshold quantile $t$, while the y-axis represents the corresponding number of clusters $k^*$ obtained by SNC, with the red dashed line indicating the ground-truth number of clusters $K$ of the data. In this test, the range of $t$ was set from 20\% to 40\%, with a step size of 2.5\%. The results show that for most datasets, a reasonable $k^*$ value could be adaptively found within this range, and when the $k^*$ obtained by SNC matched $K$, the corresponding $t$ was not a single fixed value but rather a range. This indicates that SNC is not overly sensitive to the choice of parameter $t$, exhibiting a degree of robustness. 

\begin{figure}
\centering
\includegraphics[width = 1\textwidth]{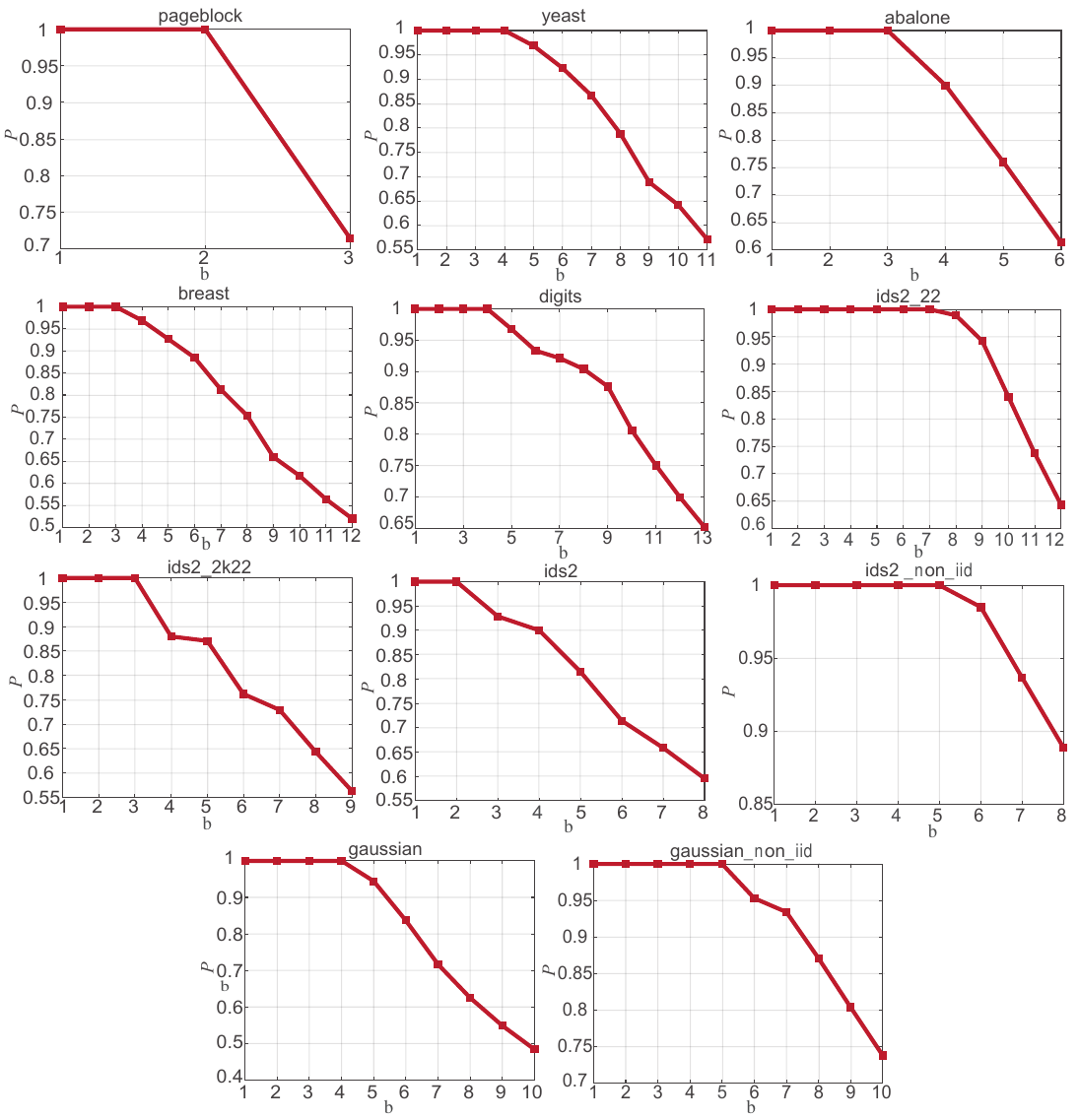}
  \caption{The variation curve of $P$ as $b$ changes on different datasets. The impact of the $b$-nearest neighbor parameter ($b$) used in the SNC algorithm on the proportion of short-distance point pairs ($P$) across different datasets. From the graph, it can be seen that the proportion of short-distance point pairs ($P$) shows a similar trend as $b$ increases across all datasets, which aligns with our expectations.}
  \label{rFk}
\end{figure}

\begin{figure}
\centering
\includegraphics[width = 1\textwidth]{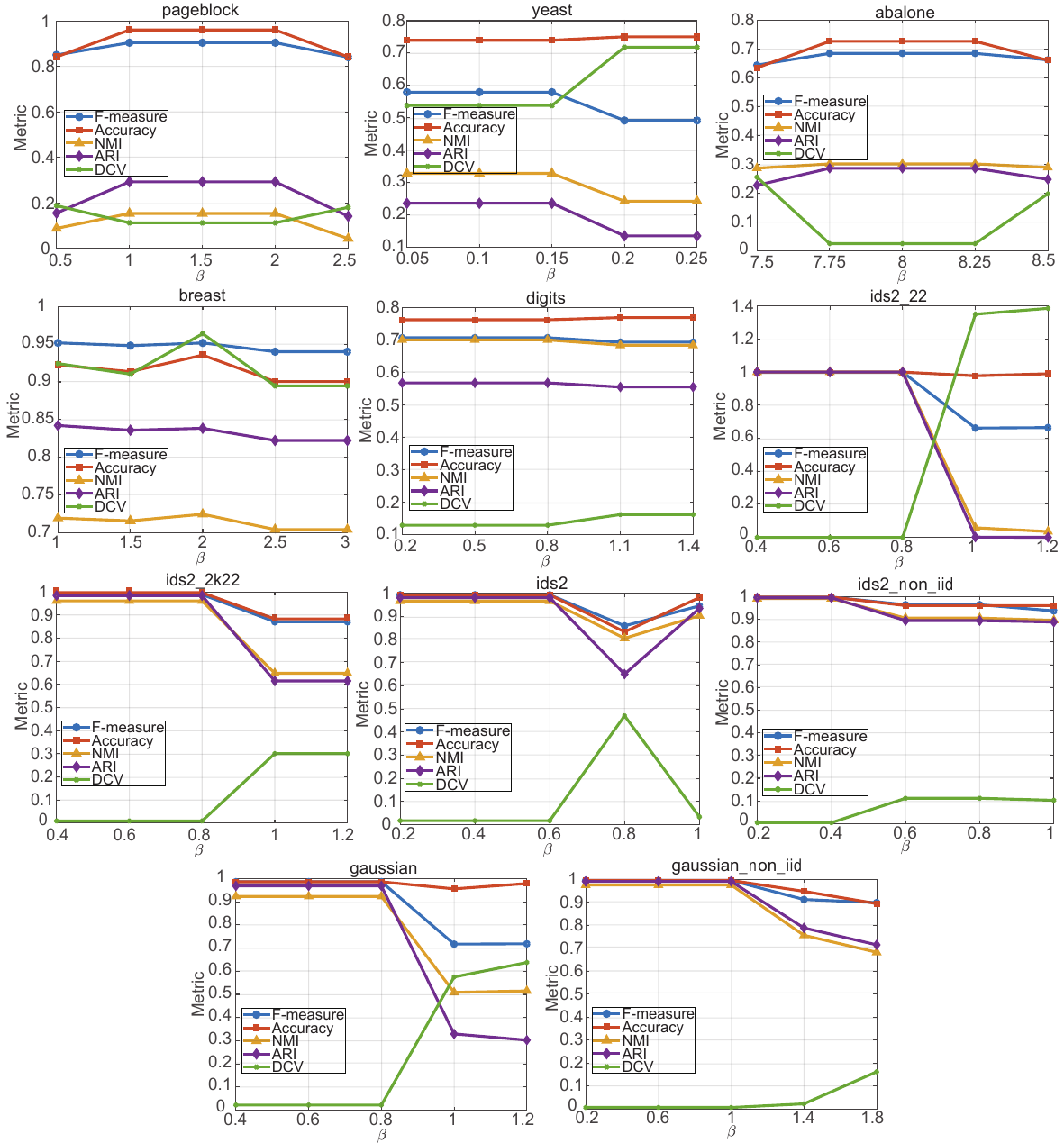}
  \caption{Parameter sensitivity analysis on $\beta$. The impact of using different $\beta$ values on the clustering results of the global clustering algorithm across each dataset was studied. The five lines represent the results of the F-measure, Accuracy, NMI, ARI, and DCV metrics, where a smaller DCV value indicates better performance, and the higher the value of the other four metrics, the better the performance.}
  \label{rWb}
\end{figure}

In addition, the impact of the $b$ parameter (used in the SNC) on the proportion of short-distance pairs ($P$) across different datasets was analyzed, as shown in Figure \ref{rFk}. In the figure, the x-axis represents various $b$ values, while the y-axis indicates the corresponding proportion of short-distance pairs. Since $b$ directly affects the determination of neighbor relationships and influences whether an appropriate number of clusters $k^*$ can be identified, selecting an optimal $b$ is important. From the results, it can be observed that the proportion of short-distance pairs across datasets shows a similar pattern with increasing $b$: initially maintaining a value of 1, then gradually declining. In the algorithm, the value of $b$ was chosen at the point where this proportion begins to decrease as the criterion for determining a suitable $b$. The results of the experiment indicate that a suitable $b$ can be identified across all datasets, aligning with our expectations of SNC's accuracy.

In the experiment of the global clustering, the impact of different $\beta$ on clustering results across each dataset was studied, as shown in Figure \ref{rWb}. The x-axis represents the values of $\beta$, and the y-axis shows the results of various evaluation metrics. Each subplot corresponds to the clustering performance for different $\beta$ values on a specific dataset. The five lines represent the results for F-measure, Accuracy, NMI, ARI, and DCV, where a smaller DCV value indicates better performance, while higher values for the other four metrics indicate better performance. According to the results, the datasets: $pageblock$, $yeast$, $abalone$, $breast$, and $ids2\_non\_iid$ achieve better clustering results when $\beta$ is set to 1.5, 0.1, 8, 2, and 0.4, respectively. The remaining six datasets perform better in the neighborhood of $\beta=0.5$. This indicates that, due to different data distributions across datasets, the sensitivity to $\beta$ also varies. However, each dataset shows improved clustering results in the neighborhood of a specific $\beta$. Therefore, selecting an appropriate $\beta$ is crucial. As previously discussed, a practical approach involves starting from the vicinity of $\beta=1$, adjusting $\beta$ progressively with a relatively large step size. Once performance begins to improve, the step size can be gradually reduced. When significantly better clustering results are observed within a certain range of $\beta$, that value can be regarded as appropriate. These findings confirm both the significance and the practicality of tuning $\beta$ according to the underlying distributional characteristics of the data.

\subsection{Computational Efficiency Comparison}
\begin{figure}[!t]
\centerline{\includegraphics[width=1\textwidth]{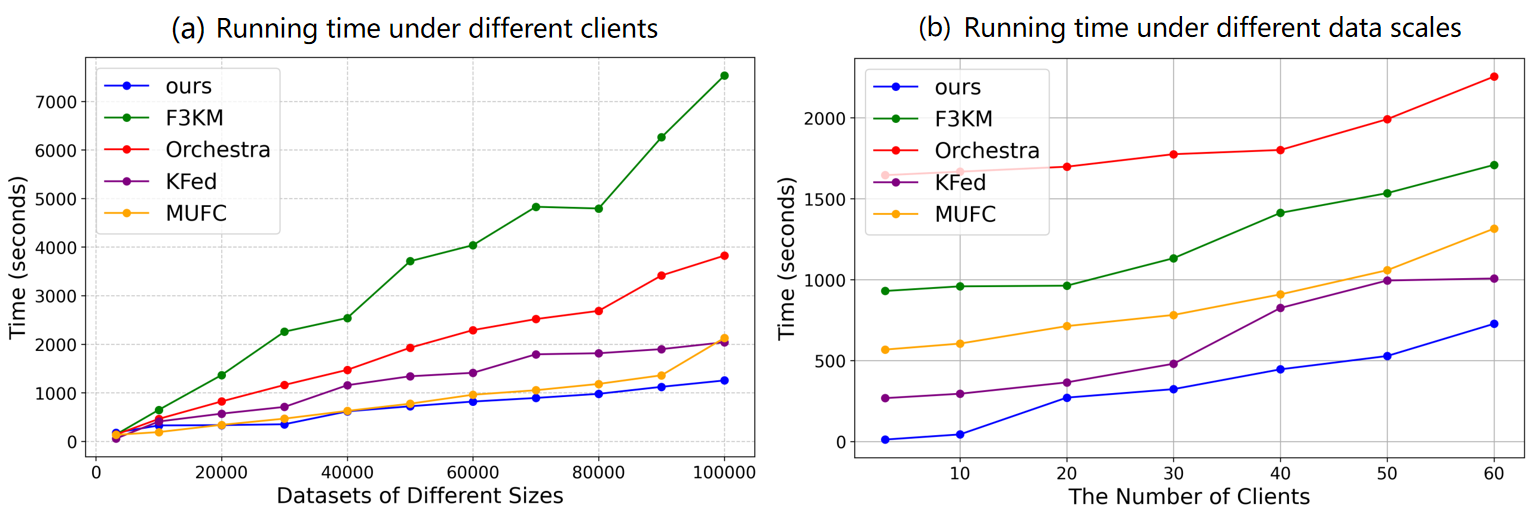}}
\caption{Time overhead analysis under two different settings: (a) varying the dataset size from {$10^3$} to {$10^5$} with 30 clients; (b) varying the number of clients from 3 to 60 with a fixed dataset size of 32,000. 
\label{Efficiency}}
\end{figure}

To assess the scalability and operational efficiency of different methods, experiments were conducted by scaling the data volume from $10^3$ to $10^5$ with the number of clients fixed at 30, and by increasing the number of clients from 3 to 60 while keeping the data volume fixed at 32,000, as shown in Figure \ref{Efficiency}. The experimental results show that the running time of each method exhibits an approximately linear increase with respect to both the number of clients and the data size. This suggests that the overall computational overhead remains relatively manageable and demonstrates the scalability of the proposed method.

Additionally, compared to other methods, Fed-$k^*$-HC demonstrates clear advantages in operational efficiency, particularly in reducing communication time. By adopting a one-shot federated clustering strategy, it completes global distribution fusion using only a single communication round, avoiding delays from multiple synchronizations. This makes Fed-$k^*$-HC well-suited for scenarios with limited communication resources.

\section{Discussions and Limitations}\label{sec5}
The proposed Fed-$k^*$-HC method significantly improves the clustering effect through fine-grained subclustering on the clients and hierarchical aggregation on the server, and also facilitates an efficient and privacy-protected one-shot communication. However, it is also not exempt from limitations, which are discussed following the key challenges of federated learning~\cite{zhang2022federated}: 1) the server efficiency under huge numbers of clients and samples; 2) the robustness in the case of extremely imbalanced cluster distributions; and 3) privacy protection standard beyond conventional federated clustering requirements, which are analyzed below: 1) \textbf{Server Efficiency:} When the number of clients is large or the sample scale is huge, the execution time on the server may rapidly increase due to the hierarchical clustering nature of the learning paradigm; 2) \textbf{Imbalance Tolerance:} The clustering performance may still fluctuate in scenarios with extremely imbalanced data distribution, as the micro-subclusters obtained on the clients can only detect small clusters to a certain degree. If the relatively small clusters are even smaller than the micro-subclusters, the proposed Fed-$k^*$-HC may be incompetent in detecting them; 3) \textbf{Privacy Protection:} Like most existing federated clustering methods, the proposed method has not yet integrated an extra privacy protection mechanism, which has also been discussed in Remark 1.

\section{Conclusion}\label{sec6}
This paper introduces a novel federated clustering framework, Fed-$k^*$-HC, to address two commonly overlooked yet prevalent challenges in practical applications: 1) the often unknown real number of clusters during the clustering process, and 2) the potential for an imbalanced global data distribution that the server must handle. To tackle these issues, a micro-partitioning strategy was devised to divide data into smaller subclusters on clients, without the need to predefine the number of clusters $k$. Based on the distribution of these subclusters, substitute data is generated and sent to the server to facilitate subsequent clustering, effectively safeguarding data privacy. On the server, the algorithm adaptively determines an appropriate number of clusters $k^*$, and performs global clustering through the hierarchical merging of subclusters. This approach effectively avoids the ``uniform effect'' seen in most federated clustering methods, showing strong performance in exploring imbalanced clusters, and eliminating the need to preset the number of clusters. The above merits of Fed-$k^*$-HC are validated through extensive experiments, including clustering performance comparison on IID, Non-IID data, balanced, and imbalanced datasets, ablation studies, hit ratio evaluation of the inferred number of clusters, hyper-parameter sensitivity studies, and efficiency evaluation. Overall, this work provides an effective federated clustering paradigm for exploring an unknown number of imbalanced clusters, which is promising in bridging the gap between the methods in this field and the real complex distributed data environment.

Despite the advancements of the proposed method, it is also not exempt from limitations as discussed in Section \ref{sec5}. Accordingly, the next avenue of this work could be to extend the proposed method to large-scale, high-dimensional, and sparse datasets by incorporating dimensionality reduction and lightweight distributed representation mechanisms. 
Furthermore, integrating differential privacy into the one-shot federated clustering framework will help balance privacy protection and clustering performance, so as to broaden the application scenarios of the proposed method.

\section*{Acknowledgements}
This work was supported in part by the National Natural Science Foundation of China under grants 62172112 and 62476063, the National Key Research and Development Program of China under grant 2022YFE0112200, the Guangdong Provincial Key Laboratory of Intellectual Property and Big Data under grant 2018B030322016, and the Natural Science Foundation of Guangdong Province under grant 2025A1515011293.

\bibliographystyle{elsarticle-num} 
\bibliography{FedkHC}

@article{konecny2016federated2,
  title="Federated learning: Strategies for improving communication efficiency",
  author="Konecn{\`y}, Jakub and McMahan, H Brendan and Yu, Felix X and Richt{\'a}rik, Peter and Suresh, Ananda Theertha and Bacon, Dave",
  journal="arXiv preprint arXiv:1610.05492",
  year="2016"}

@article{yang2019federated,
  title="Federated machine learning: Concept and applications",
  author="Yang, Qiang and Liu, Yang and Chen, Tianjian and Tong, Yongxin",
  journal="ACM Transactions on Intelligent Systems and Technology",
  volume="10",
  number="2",
  pages="1--19",
  year="2019",
  publisher="ACM New York, NY, USA"
}

@inproceedings{dennis2021heterogeneity,
  title="Heterogeneity for the win: One-shot federated clustering",
  author="Dennis, Don Kurian and Li, Tian and Smith, Virginia",
  booktitle="International Conference on Machine Learning",
  pages="2611--2620",
  year="2021",
  organization="PMLR"
}

@article{sattler2020clustered,
  title="Clustered federated learning: Model-agnostic distributed multitask optimization under privacy constraints",
  author="Sattler, Felix and M{\"u}ller, Klaus-Robert and Samek, Wojciech",
  journal="IEEE Transactions on Neural Networks and Learning Systems",
  volume="32",
  number="8",
  pages="3710--3722",
  year="2020",
  publisher="IEEE"
}

@article{ghosh2020efficient,
  title="An efficient framework for clustered federated learning",
  author="Ghosh, Avishek and Chung, Jichan and Yin, Dong and Ramchandran, Kannan",
  journal="Advances in Neural Information Processing Systems",
  volume="33",
  pages="19586--19597",
  year="2020"
}

@inproceedings{kumar2020federated,
  title="Federated k-means clustering: A novel edge ai based approach for privacy preservation",
  author="Kumar, Hemant H and Karthik, VR and Nair, Mydhili K",
  booktitle="IEEE International Conference on Cloud Computing in Emerging Markets",
  pages="52--56",
  year="2020",
  organization="IEEE"
}

@article{pedrycz2021federated,
  title="Federated FCM: clustering under privacy requirements",
  author="Pedrycz, Witold",
  journal="IEEE Transactions on Fuzzy Systems",
  volume="30",
  number="8",
  pages="3384--3388",
  year="2021",
  publisher="IEEE"
}

@article{li2022secure,
  title="Secure federated clustering",
  author="Li, Songze and Hou, Sizai and Buyukates, Baturalp and Avestimehr, Salman",
  journal="arXiv preprint arXiv:2205.15564",
  year="2022"
}

@article{xiong2009k,
  title="K-means clustering versus validation measures: A data-distribution perspective",
  author="Xiong, Hui and Wu, Junjie and Chen, Jian",
  journal="IEEE Transactions on Systems, Man, and Cybernetics, Part B: Cybernetics",
  volume="39",
  number="2",
  pages="318--331",
  year="2009",
  publisher="Institute of Electrical and Electronics Engineers Inc."
}

@article{liang2012k,
  title="The $ K $-means-type algorithms versus imbalanced data distributions",
  author="Liang, Jiye and Bai, Liang and Dang, Chuangyin and Cao, Fuyuan",
  journal="IEEE Transactions on Fuzzy Systems",
  volume="20",
  number="4",
  pages="728--745",
  year="2012",
  publisher="IEEE"
}

@article{lu2019self,
  title="Self-adaptive multiprototype-based competitive learning approach: A k-means-type algorithm for imbalanced data clustering",
  author="Lu, Yang and Cheung, Yiu-Ming and Tang, Yuan Yan",
  journal="IEEE Transactions on Cybernetics",
  volume="51",
  number="3",
  pages="1598--1612",
  year="2019",
  publisher="IEEE"
}

@misc{uci,
author = "Markelle Kelly, Rachel Longjohn, Kolby Nottingham",
title = "The UCI Machine Learning Repository",
howpublished = "\url{https://archive.ics.uci.edu}",
year = "2023"
}

@article{long2022clustering,
  title="Clustering based on local density peaks and graph cut",
  author="Long, Zhiguo and Gao, Yang and Meng, Hua and Yao, Yuqin and Li, Tianrui",
  journal="Information Sciences",
  volume="600",
  pages="263--286",
  year="2022",
  publisher="Elsevier"
}

@article{li2020federated,
  title="Federated learning: Challenges, methods, and future directions",
  author="Li, Tian and Sahu, Anit Kumar and Talwalkar, Ameet and Smith, Virginia",
  journal="IEEE Signal Processing Magazine",
  volume="37",
  number="3",
  pages="50--60",
  year="2020",
  publisher="IEEE"
}

@inproceedings{briggs2020federated,
  title="Federated learning with hierarchical clustering of local updates to improve training on non-IID data",
  author="Briggs, Christopher and Fan, Zhong and Andras, Peter",
  booktitle="International Joint Conference on Neural Networks",
  pages="1--9",
  year="2020",
  organization="IEEE"
}

@article{lin2017clustering,
  title="Clustering-based undersampling in class-imbalanced data",
  author="Lin, Wei-Chao and Tsai, Chih-Fong and Hu, Ya-Han and Jhang, Jing-Shang",
  journal="Information Sciences",
  volume="409",
  pages="17--26",
  year="2017",
  publisher="Elsevier"
}

@article{schubert2017dbscan,
  title="DBSCAN revisited, revisited: why and how you should (still) use DBSCAN",
  author="Schubert, Erich and Sander, J{\"o}rg and Ester, Martin and Kriegel, Hans Peter and Xu, Xiaowei",
  journal="ACM Transactions on Database Systems (TODS)",
  volume="42",
  number="3",
  pages="1--21",
  year="2017",
  publisher="Acm New York, NY, USA"
}

@article{frey2007clustering,
  title="Clustering by passing messages between data points",
  author="Frey, Brendan J and Dueck, Delbert",
  journal="science",
  volume="315",
  number="5814",
  pages="972--976",
  year="2007",
  publisher="American Association for the Advancement of Science"
}

@article{carreira2015review,
  title="A review of mean-shift algorithms for clustering",
  author="Carreira-Perpin{\'a}n, Miguel A",
  journal="arXiv preprint arXiv:1503.00687",
  year="2015"
}

@inproceedings{xie2023fed,
  title="Fed-SC: One-Shot Federated Subspace Clustering over High-Dimensional Data",
  author="Xie, Songjie and Wu, Youlong and Liao, Kewen and Chen, Lu and Liu, Chengfei and Shen, Haifeng and Tang, MingJian and Sun, Lu",
  booktitle="2023 IEEE 39th International Conference on Data Engineering (ICDE)",
  pages="2905--2918",
  year="2023",
  organization="IEEE"
}

@article{8423698,
  title={Learning self-growth maps for fast and accurate imbalanced streaming data clustering},
  author={Zhang, Yiqun and Feng, Sen and Wang, Pengkai and Tan, Zexi and Luo, Xiaopeng and Ji, Yuzhu and Zou, Rong and Cheung, Yiu-Ming},
  journal={IEEE Transactions on Neural Networks and Learning Systems},
  year={2025},
  volume={36},
  number={9},
  pages={16049-16061},
  publisher={IEEE}
}

@inproceedings{10.1007/978-981-99-8435-0_3,
author="Zhao, Mingjie
and Zhang, Yiqun
and Ji, Yuzhu
and Lu, Yang",
editor="Liu, Qingshan
and Wang, Hanzi
and Ma, Zhanyu
and Zheng, Weishi
and Zha, Hongbin
and Chen, Xilin
and Wang, Liang
and Ji, Rongrong",
title="Unsupervised Concept Drift Detection via Imbalanced Cluster Discriminator Learning",
booktitle="Pattern Recognition and Computer Vision. ",
year="2024",
publisher="Springer Nature Singapore",
address="Singapore",
pages="31--43", 
organization = ""
}

@inproceedings{
pan2023machine,
title={Machine Unlearning of Federated Clusters},
author={Chao Pan and Jin Sima and Saurav Prakash and Vishal Rana and Olgica Milenkovic},
booktitle={The Eleventh International Conference on Learning Representations },
year={2023}
}

@article{LU2023105714,
title = {Federated clustering for recognizing driving styles from private trajectories},
journal = {Engineering Applications of Artificial Intelligence},
volume = {118},
pages = {105714},
year = {2023},
author = {Lin Lu and Yao Lin and Yuan Wen and Jinxiong Zhu and Shengwu Xiong}
}

@article{Fritzke1994GCS,
  title={Growing cell structures--A self-organizing network for unsupervised and supervised learning},
  author={Bernd Fritzke},
  journal={Neural Networks},
  year={1994},
  volume={7},
  pages={1441-1460}
}

@article{knn,
author = {Cunningham, P\'{a}draig and Delany, Sarah Jane},
title = {k-Nearest Neighbour Classifiers - A Tutorial},
year = {2021},
issue_date = {July 2022},
publisher = {Association for Computing Machinery},
address = {New York, NY, USA},
volume = {54},
number = {6},
issn = {0360-0300}, 
journal = {ACM computing surveys (CSUR)}
}

@article{Zhu2016NN,
  title={Natural neighbor: A self-adaptive neighborhood method without parameter K},
  author={Qingsheng Zhu and Ji Feng and Jinlong Huang},
  journal={Pattern Recognit. Lett.},
  year={2016},
  volume={80},
  pages={30-36}
}

@article{NMI_65fed2d86db441a3919e495dfd0258a6,
title = "Information theoretic measures for clusterings comparison: variants, properties, normalization and correction for chance",
author = "Vinh Nguyen and Julien Epps and James Bailey",
year = "2010",
language = "English",
volume = "11",
pages = "2837 -- 2854",
journal = "Journal of Machine Learning Research",
issn = "1533-7928",
publisher = "Journal of Machine Learning Research (JMLR)",
}

@article{issuesINfl_Bhanbhro2024IssuesIF,
  title={Issues in federated learning: some experiments and preliminary results},
  author={Jamsher Bhanbhro and Simona Nistic{\`o} and Luigi Palopoli},
  journal={Scientific Reports},
  year={2024},
  volume={14}
}

@article{Zhu2023F3KMFF,
  title={F3KM: Federated, Fair, and Fast k-means},
  author={Shengkun Zhu and Quanqing Xu and Jinshan Zeng and Sheng Wang and Yuan Sun and Zhifeng Yang and Chuanhui Yang and Zhiyong Peng},
  journal={Proceedings of the ACM on Management of Data},
  year={2023},
  volume={1},
  pages={1 - 25}
}

@article{Wang2022FederatedCF,
  title={Federated Clustering for Electricity Consumption Pattern Extraction},
  author={Yi Wang and Mengshuo Jia and Ning Gao and Leandro Von Krannichfeldt and Mingyang Sun and Gabriela Hug},
  journal={IEEE Transactions on Smart Grid},
  year={2022},
  volume={13},
  pages={2425-2439}
}

@article{zhang2021survey,
  title={A survey on federated learning},
  author={Zhang, Chen and Xie, Yu and Bai, Hang and Yu, Bin and Li, Weihong and Gao, Yuan},
  journal={Knowledge-Based Systems},
  volume={216},
  pages={106775},
  year={2021},
  publisher={Elsevier}
}

@inproceedings{Lubana2022OrchestraUF,
  title={Orchestra: Unsupervised Federated Learning via Globally Consistent Clustering},
  author={Ekdeep Singh Lubana and Chi Ian Tang and Fahim Kawsar and Robert P. Dick and Akhil Mathur},
  booktitle={International Conference on Machine Learning},
  year={2022}, 
  organization={},
  pages={}
}

@article{Li2023OnTP,
  title={On the Privacy of Federated Clustering: a Cryptographic View},
  author={Qiongxiu Li and Lixia Luo},
  journal={ICASSP 2024 - 2024 IEEE International Conference on Acoustics, Speech and Signal Processing (ICASSP)},
  year={2023},
  pages={4865-4869}
}

@article{Hegde2021SoKEP,
  title={SoK: Efficient Privacy-preserving Clustering},
  author={Aditya Hegde and Helen M{\"o}llering and T. Schneider and Hossein Yalame},
  journal={Proceedings on Privacy Enhancing Technologies},
  year={2021},
  volume={2021},
  pages={225 - 248}
}

@article{abood2018survey,
  title={A survey on cryptography algorithms},
  author={Abood, Omar G and Guirguis, Shawkat K},
  journal={International Journal of Scientific and Research Publications},
  volume={8},
  number={7},
  pages={495--516},
  year={2018}
}

@article{ALHUTHAIFI2023833,
title = {Federated learning in smart cities: Privacy and security survey},
journal = {Information Sciences},
volume = {632},
pages = {833-857},
year = {2023},
issn = {0020-0255},
doi = {https://doi.org/10.1016/j.ins.2023.03.033},
author = {Rasha Al-Huthaifi and Tianrui Li and Wei Huang and Jin Gu and Chongshou Li}
}

@article{ZeinabPPFL2025,
title = {Privacy-preserving federated learning compatible with robust aggregators},
journal = {Engineering Applications of Artificial Intelligence},
volume = {143},
pages = {110078},
year = {2025},
issn = {0952-1976},
doi = {https://doi.org/10.1016/j.engappai.2025.110078},
author = {Zeinab Alebouyeh and Amir Jalaly Bidgoly}
}

@inproceedings{zhang2025asynchronous,
  title={Asynchronous Federated Clustering with Unknown Number of Clusters},
  author={Zhang, Yunfan and Zhang, Yiqun and Lu, Yang and Li, Mengke and Chen, Xi and Cheung, Yiu-ming},
  booktitle={Proceedings of the Thirty-Ninth AAAI Conference on Artificial Intelligence},
  pages={22695--22702},
  year={2025},
  organization={Association for the Advancement of Artificial Intelligence}
}

@article{koley2024critically,
  title={Critically Reckoning Spectrophotometric Detection of Asymptomatic Cyanotoxins and Faecal Contamination in Periurban Agrarian Ecosystems via Convolutional Neural Networks},
  author={Koley, Soumyajit},
  journal={Thai Journal of Science and Technology},
  volume="",
  number="",
  pages="",
  year={2024},
  publisher={Widyothayan University},
  doi={10.48048/tis.2024.8528}
}

@article{acar2018survey,
title={A survey on homomorphic encryption schemes: Theory and implementation},
author={Acar, Alptekin and Aksu, Hidayet and Uluagac, Ahmed S and Conti, Mauro},
journal={ACM Computing Surveys},
volume={51},
number={4},
pages={1--35},
year={2018}
}

@article{wei2020federated,
  title={Federated learning with differential privacy: Algorithms and performance analysis},
  author={Wei, Ke and Li, Junyi and Ding, Ming and Ma, Chang and Yang, Hua-Hua and Farokhi, Farhad and Jin, Shiqiang and Quek, Tony Q.S. and Poor, H. Vincent},
  journal={IEEE Transactions on Information Forensics and Security},
  volume={15},
  pages={3454--3469},
  year={2020}
}

@article{li2023differentially,
  title={Differentially private federated clustering over non-IID data},
  author={Li, Yin and Wang, Shusen and Chi, Chong-Yung and Quek, Tony Q.S.},
  journal={IEEE Internet of Things Journal},
  volume="",
  number="",
  pages={6705--6721},
  year={2023}
}

@article{liu2023fedet,
title={FedET: A Communication-Efficient Federated Class-Incremental Learning Framework Based on Enhanced Transformer},
author={Liu, Chenghao and Qu, Xiaoyang and Wang, Jianzong and Xiao, Jing},
journal={arXiv preprint arXiv:2306.15347},
year={2023}
}

@article{zhang2024fedgt,
  title={FedGT: Federated Node Classification with Scalable Graph Transformer},
  author={Zhang, Zaixi and Hu, Qingyong and Yu, Yang and Gao, Weibo and Liu, Qi},
  journal={arXiv preprint arXiv:2401.15203},
  year={2024}
}

@inproceedings{zhang2022federated,
title={Federated Learning Challenges and Opportunities: An Outlook},
author={Zhang, Wei and others},
booktitle={ICASSP 2022 - 2022 IEEE International Conference on Acoustics, Speech and Signal Processing (ICASSP)},
pages={9746925},
year={2022},
organization={IEEE}
}

@article{nazir2023federated,
  author = {Sajid Nazir and Mohammad Kaleem},
  title = {Federated Learning for Medical Image Analysis with Deep Neural Networks},
  journal = {Diagnostics},
  volume = {13},
  number = {9},
  pages = {1532},
  year = {2023},
  doi = {10.3390/diagnostics13091532},
}

@article{mustafa2024federated,
  author = {Mustafa Abdul Salam, Khaled M.Fouad, Doaa L.Elbably and Salah M.Elsayed},
  title = {Federated learning model for credit card fraud detection with data balancing techniques},
  journal = {Neural Computing and Applications},
  volume = {36},
  pages = {6231--6256},
  year = {2024},
  doi = {10.1007/s00521-023-09410-2},
}

\end{document}